\documentclass{article}

\usepackage[english]{babel}
\usepackage[letterpaper,top=2cm,bottom=2cm,left=3cm,right=3cm,marginparwidth=1.75cm]{geometry}

\usepackage{amsmath}
\usepackage{graphicx}
\usepackage{xurl}
\usepackage{parskip}
\usepackage{authblk}
\usepackage{physics}
\usepackage{bm}
\usepackage{todonotes}
\usepackage{natbib}
\usepackage{multirow}

\title{Gradient-Free Training of Recurrent Neural Networks using Random Perturbations}

\date{}

\author[]{Jesús García Fernández}
\author[]{Sander Keemink}
\author[]{Marcel van Gerven}
\affil[1]{Department of Machine Learning and Neural Computing, Donders Institute for Brain, Cognition and Behaviour\\
  Radboud University, Nijmegen, the Netherlands}

\begin{document}

\maketitle

\begin{abstract}
\noindent Recurrent neural networks (RNNs) hold immense potential for computations due to their Turing completeness and sequential processing capabilities, yet existing methods for their training encounter efficiency challenges. Backpropagation through time (BPTT), the prevailing method, extends the backpropagation (BP) algorithm by unrolling the RNN over time. However, this approach suffers from significant drawbacks, including the need to interleave forward and backward phases and store exact gradient information. Furthermore, BPTT has been shown to struggle to propagate gradient information for long sequences, leading to vanishing gradients. An alternative strategy to using gradient-based methods like BPTT involves stochastically approximating gradients through perturbation-based methods. This learning approach is exceptionally simple, necessitating only forward passes in the network and a global reinforcement signal as feedback. Despite its simplicity, the random nature of its updates typically leads to inefficient optimization, limiting its effectiveness in training neural networks. In this study, we present a new approach to perturbation-based learning in RNNs whose performance is competitive with BPTT, while maintaining the inherent advantages over gradient-based learning. To this end, we extend the recently introduced activity-based node perturbation (ANP) method to operate in the time domain, leading to more efficient learning and generalization. We subsequently conduct a range of experiments to validate our approach. Our results show similar performance, convergence time and scalability when compared to BPTT, strongly outperforming standard node perturbation and weight perturbation methods. These findings suggest that perturbation-based learning methods offer a versatile alternative to gradient-based methods for training RNNs which can be ideally suited for neuromorphic computing applications.
\end{abstract}

\section{Introduction}

Recurrent neural networks (RNNs), with their ability to process sequential data and capture temporal dependencies, have found applications in tasks such as natural language processing~\citep{cho2014learning, yao2013recurrent, sutskever2014sequence} and time series prediction~\citep{hewamalage2021recurrent}. They hold immense potential for computation thanks to their Turing completeness~\citep{chung2021turing}. Furthermore, due to their sequential processing capabilities, they offer high versatility to process variable-sequence length inputs and fast inference on long sequences~\citep{orvieto2023resurrecting}. Nevertheless, traditional training methods like backpropagation through time (BPTT)~\citep{werbos1990backpropagation} are challenging to apply~\citep{bengio1994learning, lillicrap2019backpropagation}, particularly with long sequences. Unrolling the RNN over time for gradient propagation and weight updating proves to be computationally demanding and difficult to parallelize with variable-length sequences. Additionally, employing BPTT can result in issues like vanishing or exploding gradients~\citep{pascanu2013difficulty}. Moreover, the non-locality of their updates can pose significant challenges when implemented on unconventional computing platforms~\citep{kaspar2021rise}.

\begin{figure}[!ht]
\centering
    \begin{flushleft} \textbf{a)}\end{flushleft}
\includegraphics[clip, trim=0cm 0cm 7.5cm 0cm, width=\textwidth]{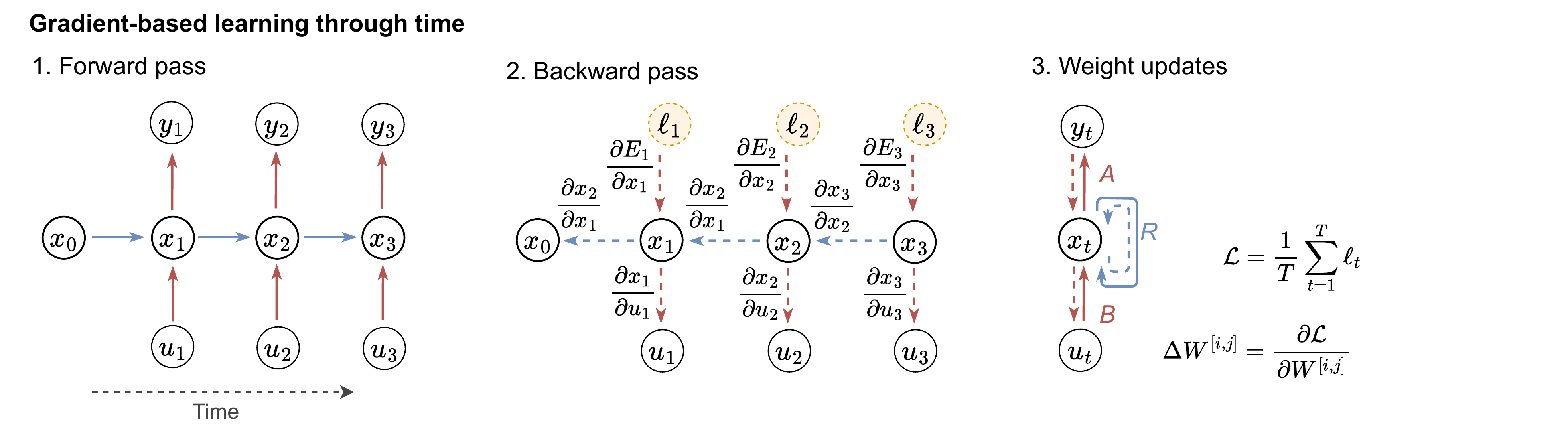}
\vspace{-20pt}
    \begin{flushleft} \textbf{b)}\end{flushleft}
\includegraphics[clip, trim=-0.5cm 0cm 5cm 0cm, width=\textwidth]{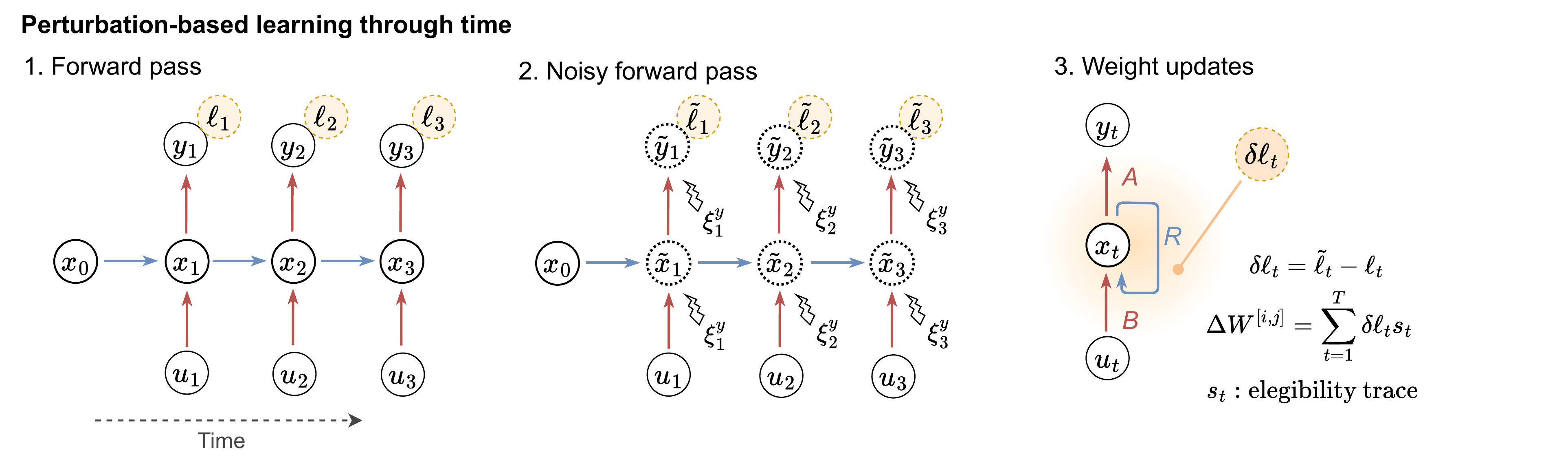}
\caption{\textbf{Gradient-based vs perturbation-based learning.} Example depicts networks unrolled across 3 time steps. \textbf{a)} General procedure followed by gradient-based learning approaches. Sequential computation of the forward and backward passes is necessary to calculate updates. \textbf{b)} General procedure utilized by perturbation-based learning approaches. The computation of the eligibility trace varies based on the employed algorithm (e.g., NP, WP, ANP). In perturbation-based learning, the forward pass and noisy forward pass can be parallelized by employing two models.}
\label{fig:main}
\end{figure}

An alternative approach to training neural networks is stochastically approximating gradients through perturbation-based methods~\citep{spall1992multivariate, widrow199030, werfel2003learning}. In this type of learning, synaptic weights are adjusted based on the impact of introducing perturbations to the network. When the perturbation enhances performance, the weights are strengthened, driven by a global reinforcement signal, and vice versa. This method is computationally simple, relying solely on forward passes and a reinforcement signal distribution across the network. It differs from gradient-based methods, such as BP, which needs a specific feedback circuit to propagate specific signals and a dedicated backward pass to compute explicit errors. This simplicity is especially beneficial for RNNs, as it eliminates the need to unroll the network over time during training~\citep{werbos1990backpropagation}. Examples of gradient-based and perturbation-based approaches to update the neural weights are visually depicted in Figure~\ref{fig:main}. Standard perturbation methods include node perturbation (NP), where the perturbations are added into the neurons, and weight perturbation (WP), where the perturbations are added into the synaptic weights~\citep{werfel2003learning, zuge2023weight}. 

Nevertheless, the stochastic nature of perturbation-based updates can lead to inefficiencies in optimizing the loss landscape~\citep{lillicrap2020backpropagation}, resulting in prolonged convergence times~\citep{werfel2003learning}. Additionally, perturbation-based often exhibit poor scalability, leading to inferior performance when compared to gradient-based methods~\citep{hiratani2022stability}. This performance gap increases as the network size increases, and in large networks, instability often arises manifesting as extreme weight growth~\citep{hiratani2022stability}.
Due to these challenges, a number of recent implementations have aimed to establish perturbation-based learning as an effective gradient-free method for neural network training. In~\citep{lansdell2019learning}, NP is utilized to approximate the feedback system in feedforward and convolutional networks, mitigating the error computation aspect of the BP algorithm. However, a dedicated feedback system is still necessary to propagate synapse-specific errors. In~\citep{zuge2023weight}, standard NP and WP are employed on temporally extended tasks in RNNs, though direct comparisons with gradient-based algorithms like BP are lacking. Their results suggest that WP may outperform NP in specific cases.

In this study, we present an implementation of perturbation-based learning in RNNs whose performance is competitive with BP, while maintaining the inherent advantages over gradient-based learning. To this end, we extend the activity-based node perturbation (ANP) approach~\citep{dalm2023effective} to operate in the time domain using RNNs. This approach relies solely on neural activities, eliminating the need for direct access to the noise process. The resulting updates align more closely with directional derivatives, compared to standard NP, approximating SGD more accurately. As a result, this approach significantly outperforms the standard NP in practical tasks. 
Furthermore, we extend standard implementations of NP and WP to operate in the time domain, using them as baselines along with BPTT, referred to as BP in this study for simplicity. In addition, we augment our time-extended implementations of ANP, NP, and WP (as well as BP) with a decorrelation mechanism, as done in the original ANP work~\citep{dalm2023effective}. This decorrelation mechanism has been shown to significantly accelerate the training of deep neural networks~\citep{dalm2024efficient} by aligning SGD updates with the natural gradient through the use of uncorrelated input variables~\citep{desjardins2015natural}. We name the resulting decorrelated methods DANP, DNP, DWP and DBP.

We assess the efficiency of our approach across various tasks. Firstly, we evaluate learning performance using three common machine learning benchmarks. Secondly, we examine the scalability of our approach to larger networks in terms of stability and task performance. Results indicate similar learning performance, convergence time, generalization and scalability compared to BP, with significant superiority over standard NP and WP. In contrast to gradient-based methods, the proposed method also offers increased versatility, with its local computations potentially rendering it compatible with neuromorphic hardware~\citep{schuman2022opportunities}. 

In the following sections, we detail our adaptation of the different perturbation-based methods utilized here, namely NP, WP and ANP. For standard approaches like NP and WP, we describe how our time-based extensions diverge from the more commonly employed temporal extensions. Additionally, we analyse the operation of the decorrelation mechanism within our RNNs. Subsequently, we describe the series of experiments conducted in this study and present the resulting outcomes.

\section{Methods}

In this section, we describe our original time-extended implementations of existing perturbation-based algorithms for feedforward networks, adapted for compatibility with RNNs and time-extended tasks. We highlight how our implementations differ from other time-extended methods in the literature. For clarity and ease of understanding, we present them in the following order: Node perturbation (NP), activity-based node perturbation (ANP), and weight perturbation (WP). These three methods will throughout be compared to backpropagation (BP). BP, WP and NP will throughout be used as baseline methods, with ANP (extended to the time-domain) being our novel contribution. Additionally, we describe the decorrelation mechanism proposed by \citet{dalm2023effective} for feedforward networks, which we incorporate into our RNN model to enhance the learning efficiency of the algorithms under consideration.

\subsection{Recurrent neural network model} \label{setup_rnn}

\begin{figure}[!ht]
\centering
\includegraphics[width=0.45\textwidth]{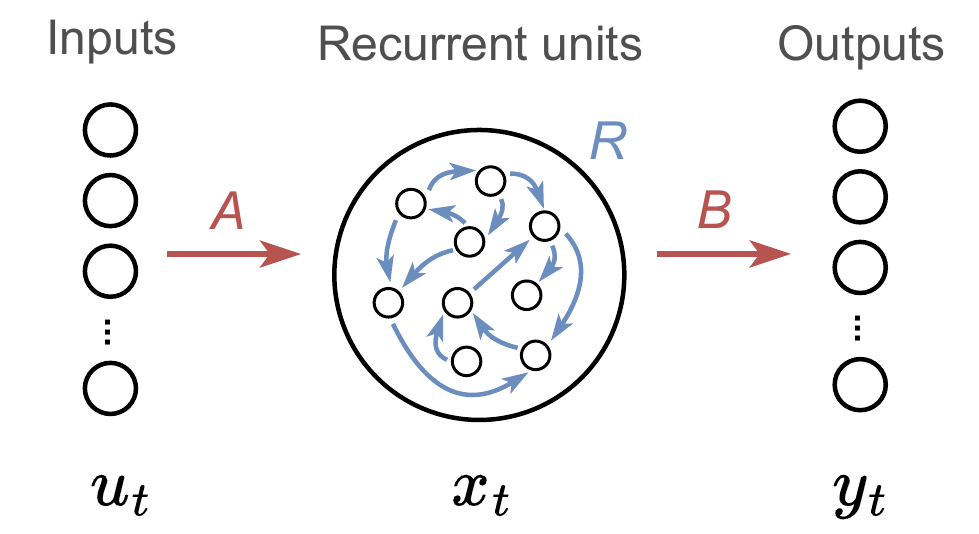}
\caption{ \textbf{Recurrent neural network model.} Recurrent units are interconnected and self-connected. Vectors $u_t$, $x_t$ and $y_t$ denote the input, recurrent and output layer activations, respectively.} 
\label{fig:rnn}
\end{figure}

In all our experiments, we employ RNNs with one hidden layer containing a large number $N$ of units with learnable weights and non-linearities in the neural outputs. The structure of the neural networks with recurrently connected hidden units is depicted in Figure~\ref{fig:rnn}, with the forward pass defined as
\begin{align*}
    x_t &= f\left(A u_t + R x_{t-1} \right) 
\\        y_t &= B x_t 
\end{align*}
where $u_t$, $x_t$ and $y_t$ denote input, hidden and output activations at time $t$, respectively, and $f(\cdot)$ represent a non-linear activation function. In our networks, we use the hyperbolic tangent activation function
$\tanh\left(x\right) = {(e^x - e^{-x})} / {(e^x + e^{-x})}$.
The parameters of the network are given by the input weights $A$, the output weights $B$ and the recurrent weights $R$.

The output error at each time step is defined as
$
    \ell_t = || y_t - y^*_t||^2 
$
where $y_t^*$ denotes the target output at time $t$. The loss associated with output $y = (y_1,\ldots,y_T)$, with $T$ the time horizon, is defined as the mean of the output errors over the entire sequence:
\begin{align*}
    \mathcal{L} &= \frac{1}{T} \sum_{t=1}^T  \ell_t \,.
\end{align*}
During learning, the network's weights undergo incremental and iterative updates to minimize the loss according to 
$
    W \leftarrow W - \eta \Delta W
$ 
with $W \in \{A, R, B\}$ the input, output and recurrent weight matrices, respectively, and $\Delta W$ their corresponding updates.
Here, $\eta$ denotes the learning rate for both forward and recurrent weights. 

The updates can be computed using different learning algorithms. In the case of backpropagation through time, the updates are given by the gradient $\nabla \mathcal{L}$ of the loss with respect to the parameters, averaged over multiple trajectories. 
Below, we describe the different perturbation-based learning methods considered in this study. 
Additionally, we describe the employed decorrelation mechanism and how it is incorporated into the networks.

\subsection{Node perturbation through time}
The node perturbation approach involves two forward passes: A standard forward pass and a noisy pass. These passes can take place either concurrently or sequentially, and the loss is computed afterwards. The weights are updated in the direction of the noise if the loss decreases, and in the opposite direction if the loss increases. During the noisy pass, noise is added to the pre-activation of each neuron as follows:
\begin{align*}
    \tilde x_t &= f\left(A u_t + R \tilde x_{t-1} + \xi_t\right) 
\\    \tilde y_t &= B \tilde x_t + \nu_t
\end{align*}
where $\tilde x_t$ and $\tilde y_t$ denote the noisy neural outputs at time $t$. The noise added in the hidden and output layers is generated from zero-mean, uncorrelated Gaussian random variables $\xi_t \sim \mathcal{N}(0, \sigma^2 I_x)$ and $\nu_t \sim \mathcal{N}(0, \sigma^2 I_y)$, respectively, where $I_x$ and $I_y$ are identity matrices with the dimensions of the hidden and output layers. The noise injected is different in every timestep $t$.

In typical implementations of node perturbation in the time domain, as seen in works like~\citep{zuge2023weight, fiete2006gradient}, the reinforcement signal is derived from the difference in loss, $\mathcal{L} = \tilde{\mathcal{L}} - \mathcal{L}$, where $\tilde{\mathcal{L}}$ and $\mathcal{L}$ represent the loss of the noisy pass and the clean pass, respectively. This signal captures the network's overall performance across the entire sequence. Additionally, an eligibility trace, computed as the sum over time of the pre-synaptic neuron's output multiplied by the injected perturbation ($\xi_t x_t^\top$, $\nu_t x_t^\top$ or $\xi_t x_{t-1}^\top$, depending on the synaptic weights being updated), is utilized. According to this method, the learning signals for updating the weights are defined in Equation \ref{eq:np_conventional}.
\begin{align}
    \Delta A &= \sigma^{-2} \delta \mathcal{L} \sum_{t=1}^T \xi_t x^{\top}_t,
\quad     \Delta B = \sigma^{-2} \delta \mathcal{L} \sum_{t=1}^T \nu_t u^{\top}_t,
\quad    \Delta R = \sigma^{-2} \delta \mathcal{L} \sum_{t=1}^T \xi_t x^{\top}_{t-1}.
\label{eq:np_conventional}
\end{align}
While this approach offers benefits such as enhanced compatibility with delayed rewards, we here consider a local approach, where we only consider the local loss difference 
$
    \delta \ell_t = \tilde \ell_t - \ell_t
$ 
in individual timesteps when computing updates in contrast to using the total loss difference $\mathcal{L}$.
This reward-per-time step is employed alongside eligibility traces local on time defined as the product of the pre-synaptic neuron's output and the injected perturbation at each time step ($\xi_t x_t^\top$, $\nu_t x_t^\top$ or $\xi_t x_{t-1}^\top$, depending on the synaptic weights being updated). 
According to our approach, the learning signals to update the weights over time are defined in Equation \ref{eq:np_ours}.
\begin{align}
    \Delta A &= \sigma^{-2} \sum_{t=1}^T \delta \ell_t \xi_t x^{\top}_t,
\quad    \Delta B = \sigma^{-2} \sum_{t=1}^T  \delta \ell_t \nu_t u^{\top}_t,
\quad    \Delta R = \sigma^{-2} \sum_{t=1}^T \delta \ell_t \xi_t x^{\top}_{t-1}.
\label{eq:np_ours}
\end{align}
Here, both the clean standard pass and the noisy pass could run concurrently using two identical copies of the model, which enables compatibility with online learning setups. This technique allows our method to compute and implement updates online at every time step.

\citet{zenke2020brain} investigate setups similar to the one proposed here, seeking to bridge the real-time recurrent learning (RTRL) algorithm~\citep{williams1989learning}, which is more effective for online setups than BPTT but computationally demanding, with biologically plausible learning rules. They demonstrate that by combining learning algorithms that approximate RTRL with temporally local losses, effective approximations can be achieved. These approximations notably decrease RTRL's computational cost while preserving strong learning performance. A similar rationale is used by~\citet{Bellec2019} in the context of recurrent spiking neural networks.
We provide a comparison between the conventional implementation and our implementation of NP through time in Appendix \ref{appendix:typicalNP_vs_ourNP}.

\subsection{Activity-based node perturbation through time}
Activity-based node perturbation (ANP) is a variant of the node perturbation approach, proposed by~\citet{dalm2023effective}, which has been exclusively applied in feedforward networks. This approach approximates the directional derivatives across the network, resulting in a closer alignment between the updates generated by this method and those provided by BP (also see~\citep{Baydin2022}). Additionally, it does not require direct access to the noise process itself as it operates solely by measuring changes in neural activity. Given that NP can be interpreted as a noisy variant of SGD~\citep{hiratani2022stability}, ANP can be seen as a more precise approximation of SGD. For a detailed derivation of the link between ANP and SGD, we refer to~\citep{dalm2023effective}. 

Similar to the node perturbation approach, the noisy pass is the same as the one performed in the standard node perturbation approach. Consistent with our node perturbation implementation extended over time, we calculate reinforcement signals at each time step to drive synaptic changes. 
Let $\alpha_t = A u_t + R x_{t-1}$ and $\beta_t = B x_t$ are the pre-activations in the clean pass. Similarly, let $\tilde \alpha_t = A u_t + R \tilde x_{t-1} + \xi_t$ and $\tilde \beta_t = B \tilde x_t + \nu_t$ denote the pre-activations in the noisy pass. Define $N$ as the total number of neurons in the network. 
We compute the learning signals responsible for weight updating as defined in Equation \ref{eq:anp_eq}.
\begin{align}
    \Delta A &= N\sum_{t=1}^T  \ell_t \frac{\delta \alpha_t}{|| \delta \alpha_t||^2} x^{\top}_t,
\quad    \Delta B = N\sum_{t=1}^T \ell_t \frac{\delta \beta_t}{||\delta \beta_t||^2} u^{\top}_t,
\quad    \Delta R = N\sum_{t=1}^T \ell_t \frac{\delta \alpha_t}{||\delta \alpha_t||^2} x^{\top}_{t-1},
\label{eq:anp_eq}
\end{align}
where $\delta \alpha_t = \tilde \alpha_t - \alpha_t$ and $\delta \beta_t = \tilde \beta_t - \beta_t$ are the pre-activation differences between the forward passes. 

\subsection{Weight perturbation through time}
Weight perturbation is an approach akin to node perturbation, where noise is injected in a second forward pass, and adjustments to the weights are made based on the resulting increase or decrease in loss. The key distinction lies in the injection of noise into the weights rather than the neural pre-activation. The noisy pass is defined as
\begin{align*}
    \tilde x_t &= f\left( \left(A + \xi_t\right) u_t + \left(R + \nu_t\right) \tilde x_{t-1} \right)
\\    \tilde y_t &= \left(B  + \zeta_t \right) \tilde x_t
\end{align*}
where $\tilde x_t$ and $\tilde y_t$ denote the noisy neural outputs at time $t$. As in node perturbation, the noise is denoted by the zero-mean, uncorrelated Gaussian random variables $\xi_t \sim \mathcal{N}(0, \sigma^2 I_{A})$, $\nu_t \sim \mathcal{N}(0, \sigma^2 I_{R})$ and $\zeta_t \sim \mathcal{N}(0, \sigma^2 I_{B})$, where $I_{A}$, $I_R$ and $I_{B}$ are identity matrices with the dimensions of $A$, $R$ and $B$, with distinct values for each timestep $t$. 

Similar to node perturbation in the time domain, typical implementations of weight perturbation in the time domain, such as those described by~\citep{zuge2023weight, cauwenberghs1992fast}, derive the reinforcement signal from the difference in loss $\delta \mathcal{L} = \tilde{\mathcal{L}} - \mathcal{L}$, computed over the entire sequence. According to this method, the learning signals for updating the weights are defined as in Equation \ref{eq:wp_conventional}.
\begin{align}
    \Delta A &= \sigma^{-2} \delta \mathcal{L} \xi_t, 
\quad    \Delta B = \sigma^{-2} \delta \mathcal{L} \zeta_t, 
\quad    \Delta R = \sigma^{-2} \delta \mathcal{L} \nu_t.
\label{eq:wp_conventional}
\end{align}
This approach inherits the same set of drawbacks and benefits as seen in node perturbation. Hence, we again employ reinforcement signals, $\delta \ell_t$, computed at each time step, to drive synaptic changes. We define the learning signals to update the weights in the time domain as in Equation \ref{eq:wp_ours}.
\begin{align}
    \Delta A &= \sigma^{-2} \sum_{t=1}^T \delta \ell_t \xi_t
\,,\quad    \Delta B = \sigma^{-2} \sum_{t=1}^T \delta \ell_t \zeta_t 
\,,\quad    \Delta R = \sigma^{-2} \sum_{t=1}^T \delta \ell_t \nu_t  \,.
\label{eq:wp_ours}
\end{align}

\subsection{Decorrelation of neural inputs}

Decorrelating neural input allows for more efficient neural representation by reducing the redundancy in neural activity, leading to improved efficiency in learning and faster learning rates. This phenomenon has found support in both biological studies~\citep{wiechert2010mechanisms, cayco2017sparse} and artificial neural network research~\citep{desjardins2015natural, huang2018decorrelated, luo2017learning, ahmad2022constrained}. It has also been shown that it significantly accelerates the training of deep neural networks~\citep{dalm2024efficient} by aligning SGD updates with the natural gradient through the use of uncorrelated input variables~\citep{desjardins2015natural}.
Additionally, in the context of global reinforcement methods such as WP or NP, where weight updates introduce substantial noise, the addition of decorrelation proves beneficial as it makes neural networks less sensitive to noise~\citep{tetzlaff2012decorrelation}. 

\begin{figure}[!ht]
\centering
\includegraphics[width=0.6\textwidth]{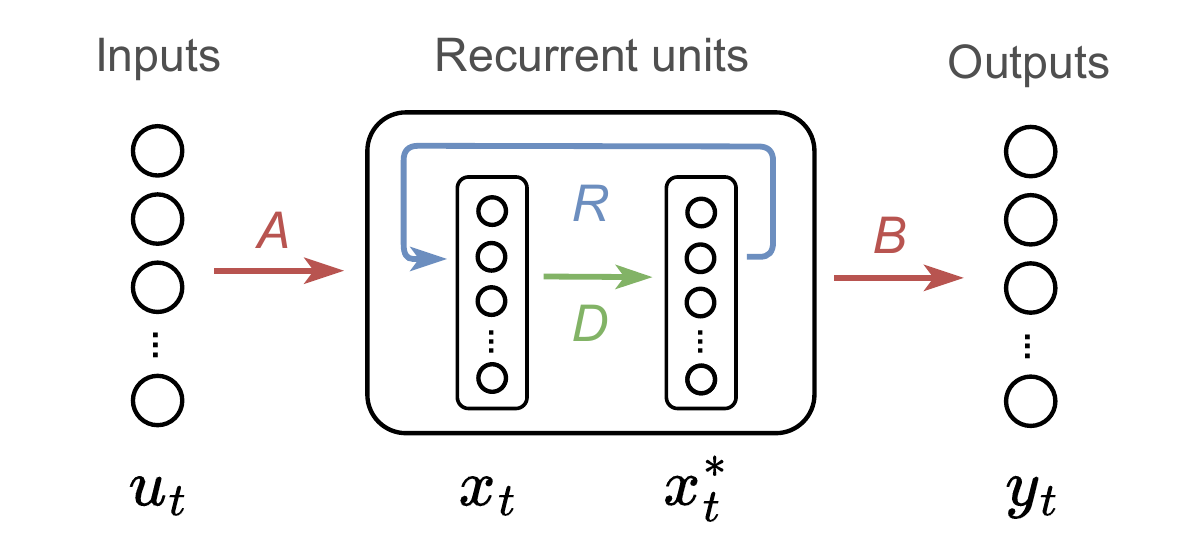}
\caption{\textbf{RNN with decorrelation scheme.} 
In this setup, we include an extra matrix, $D$, and an intermediate state that transforms the correlated neural input $x_t$, in  uncorrelated neural input $x^*_t$. The recurrent connection, $R$ is placed after the decorrelated state $x^*_t$ feeding an input to $x_{t+1}$ (in the next time step). The recurrent connection $R$ is fully connected. $x_t$ is the only variable that includes non-linearities, $u_t$, $x^*_t$ and $y_t$ are linear. The variable $x^*_t$ is used to map the recurrent states to the outputs. }
\label{fig:drnn}
\end{figure}

Given these benefits, we investigate the integration of a decorrelation mechanism into our networks, presented in~\citet{dalm2023effective}. We apply this mechanism solely to the hidden units, as the inputs fed into the network are low-dimensional, thus decorrelating them would not yield any noticeable changes and would increase computational costs. 
This involves the transformation of correlated hidden layer input $x_{t-1}$ into decorrelated input $x^*_{t-1}$ via a linear transform $x^*_{t-1} = D x_{t-1}$ with $D$ the decorrelation matrix. 
The update of recurrent units in the network is in this case defined as
$
    x_t = f\left(A u_t + R x^*_{t-1}\right)
$
and the resulting neural connectivity is depicted in Figure~\ref{fig:drnn}.

The updates of the decorrelation weights are performed using a particularly efficient learning rule proposed by~\citet{ahmad2022constrained}, which aims at reducing the cross-correlation between the neural outputs. The update is given by
$
    D \leftarrow D - \epsilon \Delta D
$
with learning rate $\epsilon$ and update defined in Equation \ref{eq:decorrelation_eq}.
\begin{equation}
   \Delta D =  \left(x^*_t \left(x^*_t \right)^\top - \text{diag}\left(\left(x^*_t \right)^2\right)\right)D
\label{eq:decorrelation_eq}
\end{equation}
for $1 \leq t \leq T$. 
The updates of the decorrelation weights are performed in an unsupervised manner, in parallel with  learning of the forward weights. We use DANP to refer to the combination of decorrelation with activity-based node perturbation.

\subsection{Experimental validation}

We evaluate the effectiveness of the described methods in training RNNs using a series of experiments encompassing several objectives. Firstly, we evaluate the performance of the networks using three standard machine learning benchmarks: Mackey-Glass time series prediction, the copying memory task and a weather prediction task. These tasks are commonly employed in the literature to evaluate the performance of RNNs and other time-series prediction models. Secondly, we assess the scalability of the considered networks when incorporating an increased number of units. Lastly, we investigate in more detail the functioning of the decorrelation mechanism. Five different runs with random seeds are carried out for each experiment. The averages of these runs are then depicted with error bars indicating maximum and minimum values.

Throughout these evaluations, we assess the perturbation-based learning methods NP, WP, and ANP, alongside the gradient-based BP learning method in conjunction with the Adam optimizer~\citep{kingma2014adam} for comparative analysis. 
Please note that for simplicity, we use the term BP in this manuscript to refer to both backpropagation and its time-domain application, backpropagation through time (BPTT)~\citep{werbos1990backpropagation}, as they are essentially the same algorithm, with the distinction being the unrolling of the network over time. Similarly, the terms NP, WP, and ANP are used in this context to refer to their application in the time domain.
Subsequently, we enhance all the methods by incorporating the decorrelation mechanism previously described into the hidden units of our networks. The resulting extended methods are named decorrelated node perturbation (DNP), decorrelated weight perturbation (DWP), decorrelated activity-based node perturbation (DANP) and decorrelated backpropagation (DBP). Detailed hyperparameters for each experiment can be found in Appendix \ref{appendix:hyperparameters}. 

Our networks and experimental setups are developed using the programming language Python~\citep{van1995python} and the deep learning framework PyTorch~\citep{paszke2019pytorch}. The BPTT algorithm is implemented using the automatic differentiation tool from PyTorch. For reproducibility and to access the implementation code, please visit our GitHub repository \footnote{ \url{https://github.com/jesusgf96/NP_RNNs}.}.

\section{Results}


\subsection{Mackey-Glass time series task}

The Mackey-Glass time series task is a classic benchmark for assessing the ability of neural networks to capture and predict chaotic dynamical systems. The data is a sequence of one-dimensional observations generated using the Mackey-Glass delay differential equations~\citep{mackey1977oscillation}, resulting in a nonlinear, delayed, and chaotic time series. We reproduce the setup of~\citet{voelker2019legendre}, where the model is tasked with predicting 15 time steps into the future with a time constant of 17 steps and a sequence length of 5000 time steps.

    \begin{figure}[!ht]
    \centering
    \begin{flushleft}\:\: \:\: \:\: \:\: \textbf{a)}\end{flushleft} 
    \includegraphics[width=0.87\textwidth]{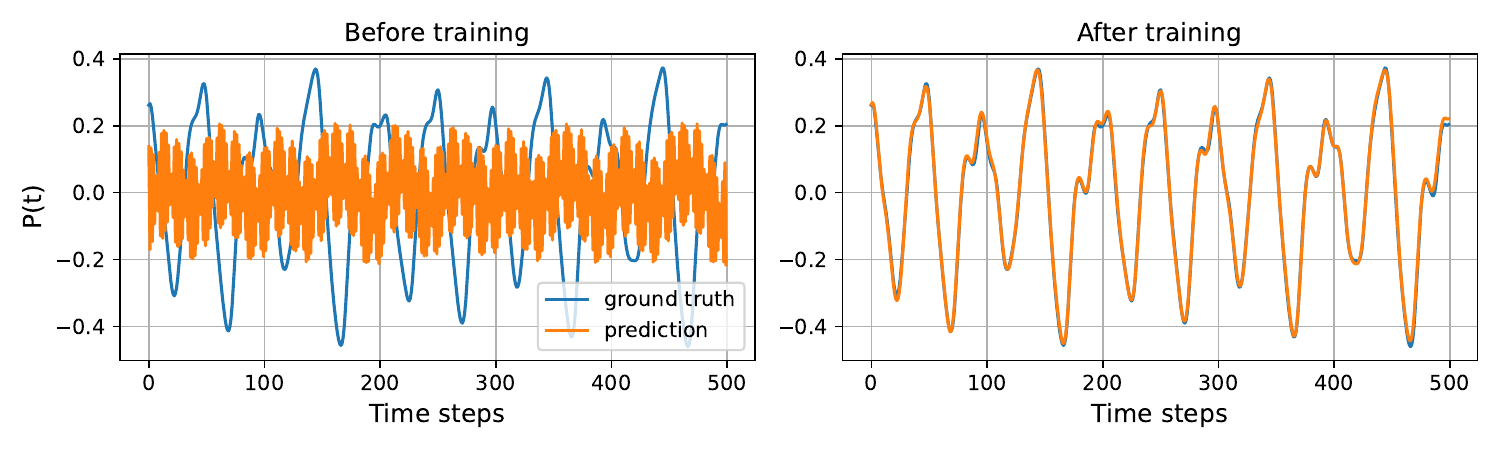}
    \vspace{-20pt}
    \begin{flushleft}\:\: \:\: \:\: \:\: \textbf{b)}\end{flushleft}
    \includegraphics[width=0.87\textwidth]{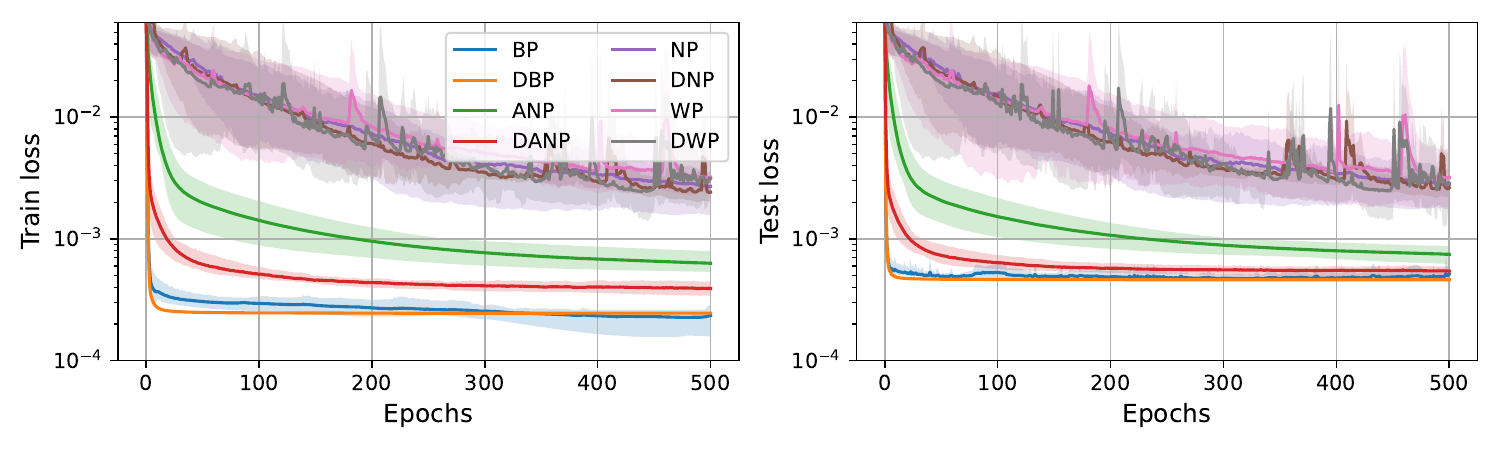}
    \vspace{-20pt}
    \begin{flushleft}\:\: \:\: \:\: \:\: \textbf{c)}\end{flushleft}
    \includegraphics[width=0.87\textwidth]{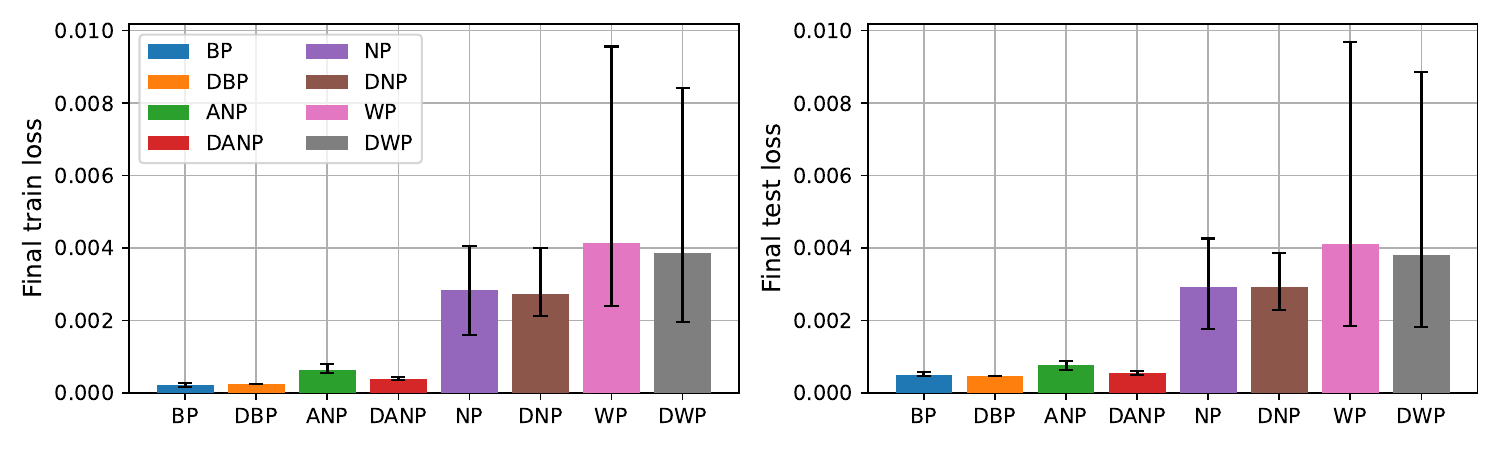}
    \caption{\textbf{Mackey-Glass data and results.} \textbf{a)} 500 time steps of a synthetically generated Mackey-Glass time series along with the predictions of a BP-trained model before and after training. \textbf{b)} Performance during training over the train and test set for the different methods, represented in a logarithmic scale. \textbf{c)} Final performance for the different methods, computed as the mean performance over the last 50 epochs.}
    \label{fig:MG}
    \end{figure}

Figure~\ref{fig:MG} depicts the results for the Mackey-Glass experiment.
Figure~\ref{fig:MG}a visualizes 500 time steps of a synthetically generated Mackey-Glass time series. Additionally, we depict the predictions made by a BP-trained model both before and after training. This example provides a visual understanding of the dataset used in our study.
Figure~\ref{fig:MG}b shows the performance during training over the train and test set for the different methods. Figure~\ref{fig:MG}c shows the final performance, computed as the mean performance over the last 50 epochs, facilitating a quantitative comparison between methods. 

The outcomes of this experiment reveal that standard perturbation-based methods NP and WP, adapted to operate in the time domain, exhibit significantly inferior performance compared to the BP baseline. The convergence time and final performance of ANP, in contrast, closely approach those of the BP baseline, especially when augmented with the decorrelation mechanism. 
Introducing the decorrelation mechanism to BP does not appear to result in a pronounced difference in convergence and final performance.

\subsection{Copying memory task}
\label{copying_memory_task}

The copying memory task is another well-established task for evaluating the memorization capabilities of recurrent models/units. In this task, the model must memorize a sequence of bits and return the exact same sequence after a specified delay period. We build on the setup of~\citet{arjovsky2016unitary}, where the sequence comprises 8 distinct bit values, with 1 extra bit serving to mark the delay period and another extra bit indicating to the network when to reproduce the sequence. In our experiments, we use a length sequence of 100 and a delay period equal to 10.

    \begin{figure}[!ht]
    \centering
    \begin{flushleft}\:\: \:\: \:\: \:\: \textbf{a)}\end{flushleft} 
    \includegraphics[width=0.8\textwidth]{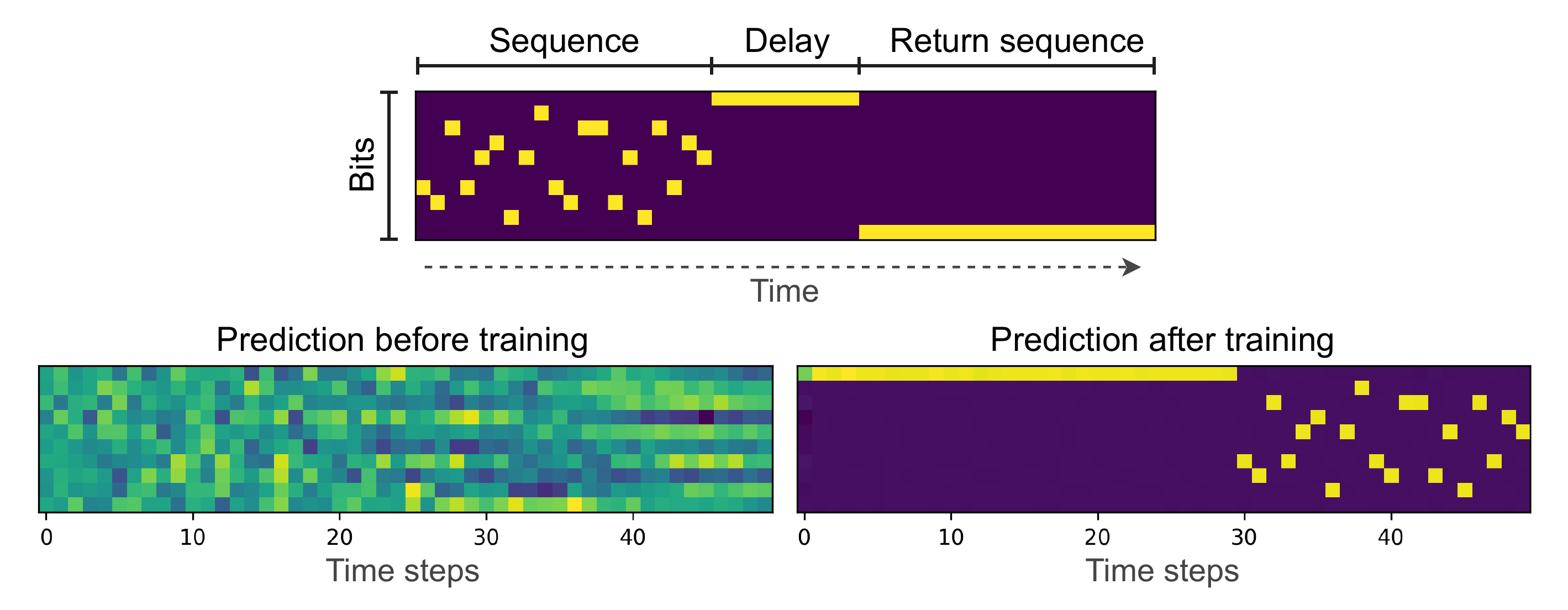}
    \vspace{-20pt}
    \begin{flushleft}\:\: \:\: \:\: \:\: \textbf{b)}\end{flushleft} 
    \includegraphics[width=0.87\textwidth]{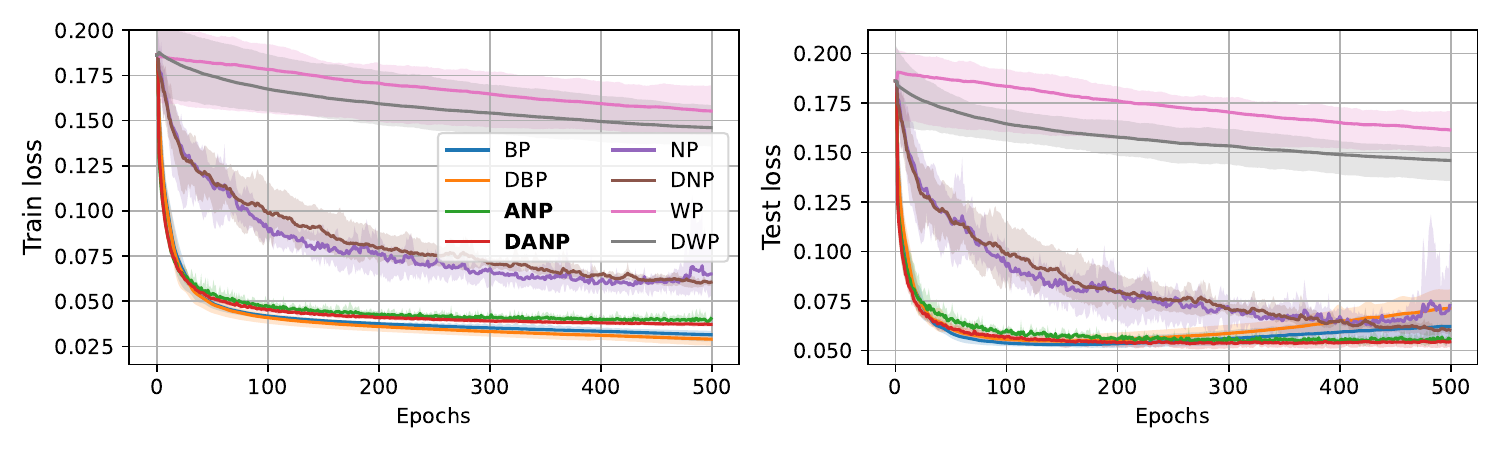}
    \vspace{-20pt}
    \begin{flushleft}\:\: \:\: \:\: \:\: \textbf{c)}\end{flushleft} 
    \includegraphics[width=0.87\textwidth]{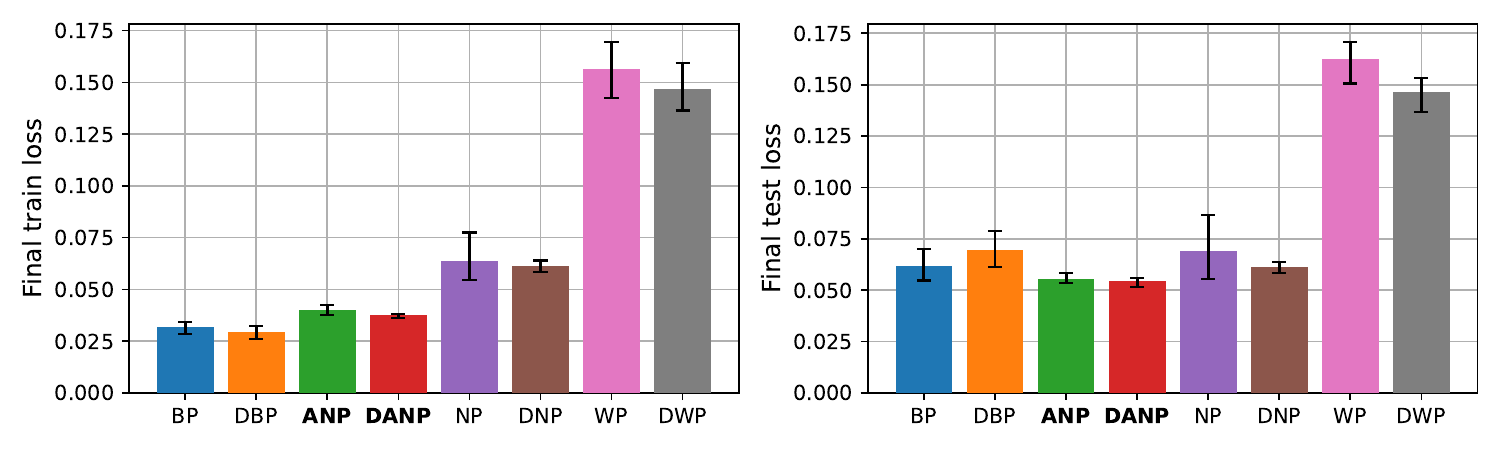}
    \caption{\textbf{Copying memory data and results.} \textbf{a)} At the top, we depict an example of an input with annotations. The sequence length is 20 and the delay period is 10. At the bottom, we show the predictions of a BP-trained model before and after training. \textbf{b)} Performance during training over the train and test set for the different methods. \textbf{c)} Final performance for the different methods, computed as the mean performance over the last 50 epochs.}
    \label{fig:copying}
    \end{figure}

Figure~\ref{fig:copying} depicts the results for the copying memory task. 
In Figure~\ref{fig:copying}a, we present a visualization of a sample of synthetically generated data for this task along with annotations. Additionally, we depict the predictions made by a BP-trained model both before and after training. This example serves to provide a visual understanding of the dataset used in our study.
In Figure~\ref{fig:copying}b, we present the performance during training over the train and test sets. Figure~\ref{fig:copying}c shows the final performance, computed as the mean performance over the last 50 epochs, facilitating a quantitative comparison between methods. 

The outcomes of this experiment reveal that standard perturbation-based methods NP and WP, adapted to operate in the time domain, exhibit significantly inferior performance compared to the BP baseline. The convergence time and final performance of ANP closely approach those of the BP baseline. In this experiment, the addition of the decorrelation mechanism does not lead to a pronounced difference in convergence and final performance, in both BP and ANP.

\subsection{Weather prediction task} \label{weather_task}

In contrast to the other benchmarks used in this paper, this task relies on real-world rather than synthetically generated data. The dataset used in this task contains climatological data spanning 1,600 U.S. locations from 2010 to 2013.\footnote{Data can be obtained from \url{https://www.ncei.noaa.gov/data/local-climatological-data/}.} We build on the setup of~\citet{zhou2021informer}, where each data point consists of one single target value to be predicted 1, 24 and 48 hours in advance, and various input climate features. In our specific configuration, we exclude duplicated features with different units of measurement (retaining Fahrenheit-measured features) and select the `Dry bulb' feature, which is a synonym for air temperature, as the target variable. The training data encompasses the initial 28 months, while the last 2 months are used for testing the model's performance. 

    \begin{figure}[!ht]
    \centering
    \begin{flushleft}\:\: \:\: \:\: \:\: \textbf{a)}\end{flushleft} 
    \includegraphics[width=0.87\textwidth]{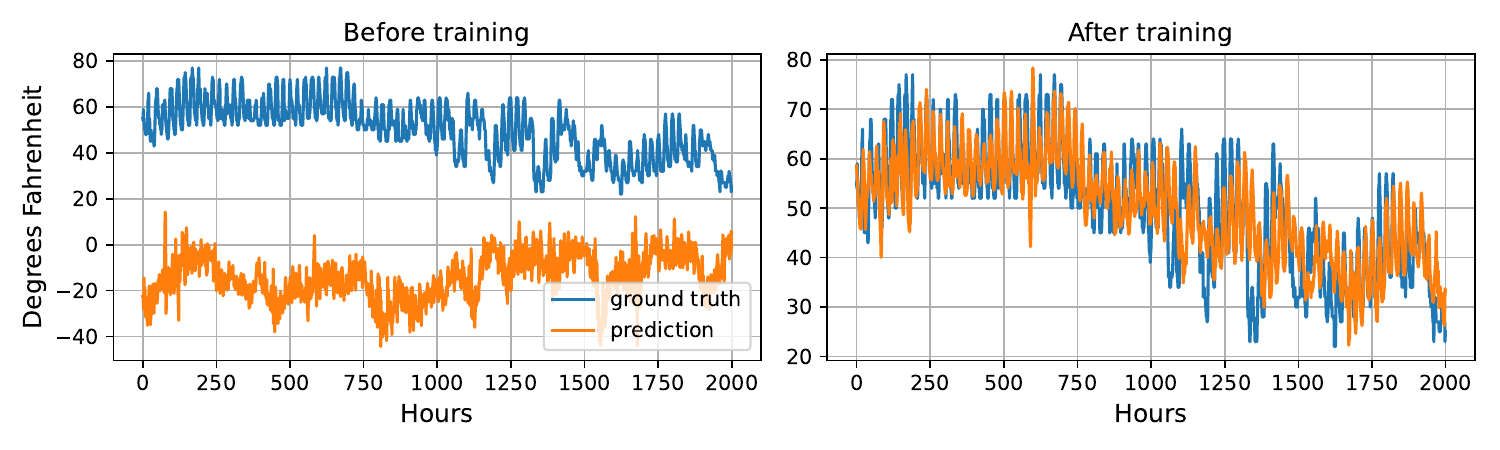}
    \vspace{-20pt}
    \begin{flushleft}\:\: \:\: \:\: \:\: \textbf{b)}\end{flushleft} 
    \includegraphics[width=0.87\textwidth]{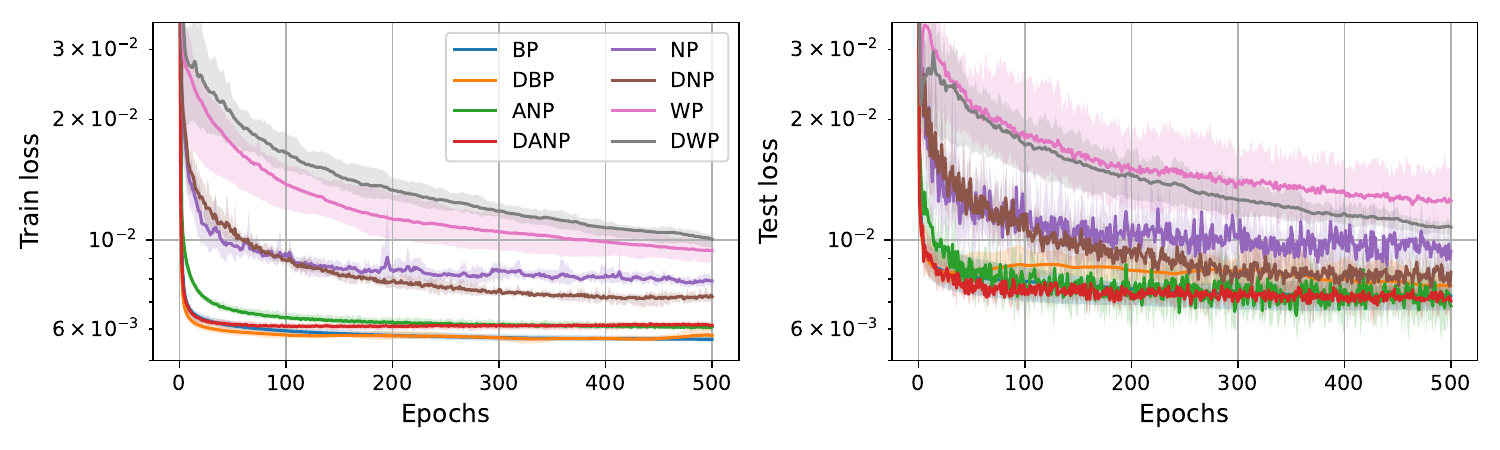}
    \vspace{-20pt}
    \begin{flushleft}\:\: \:\: \:\: \:\: \textbf{c)}\end{flushleft} 
    \includegraphics[width=0.87\textwidth]{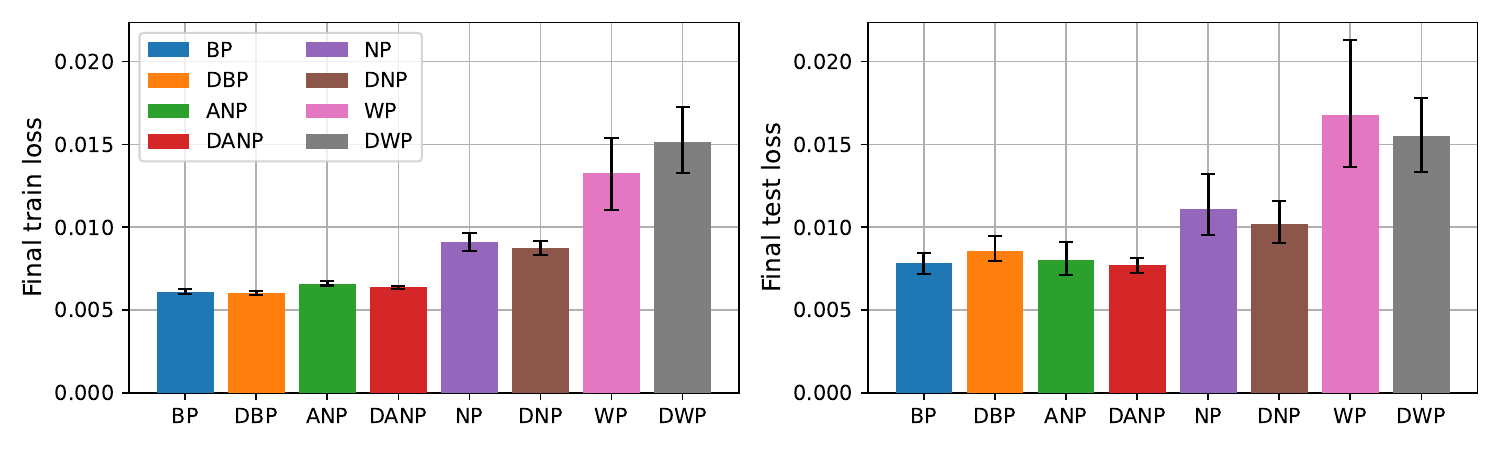}
    \caption{\textbf{48-hours ahead weather prediction data and results.} \textbf{a)} 2000 time steps of the target feature, `Dry bulb', from the weather dataset along with the predictions 48 hours ahead of a BP-trained model before and after training. \textbf{b)} Performance during training over the train and test set for the different methods, represented in a logarithmic scale. \textbf{c)} Final performance for the different methods, computed as the mean performance over the last 50 epochs.}
    \label{fig:48h_weather}
    \end{figure}

Figure~\ref{fig:48h_weather} depicts the results for this task. 
In Figure~\ref{fig:48h_weather}a, we present a visualization of 2000 time steps of the target feature, `Dry bulb', from the weather dataset. Additionally, we depict the predictions 48 hours ahead made by a BP-trained model both before and after training. This example serves to provide a visual understanding of the dataset used in our study.
In Figure~\ref{fig:48h_weather}b we present the performance during training over the train and test set for the different methods for 48-hours ahead prediction. Additionally, these figures present the final performance, computed as the mean performance over the last 50 epochs, facilitating a quantitative comparison between methods. We also include the results for 1-hour ahead and 24-hours ahead predictions in Appendix \ref{appendix:extra_exp}.

The outcomes of this experiment reveal that standard perturbation-based methods NP and WP, adapted to operate in the time domain, exhibit significantly inferior performance compared to the BP baseline. The convergence time and final performance of ANP closely approach those of the BP baseline, especially when augmented with the decorrelation mechanism, making them comparable in terms of generalization. On the contrary, introducing the decorrelation mechanism to BP results in faster convergence but compromises generalization performance.

\subsection{Scaling performance}
Here, we investigate the scalability of the described methods in RNNs with a single hidden layer and an increasingly larger number of units. We accomplish this by analyzing the final performance on the 1-hour ahead weather prediction task for networks with differing numbers of hidden units, trained using various methods. We chose this dataset as it is the most challenging among the considered datasets in this paper. The number of hidden units ranges from 100 to 3000. 

    \begin{figure}[!ht]
    \centering
    \includegraphics[width=0.87\textwidth]{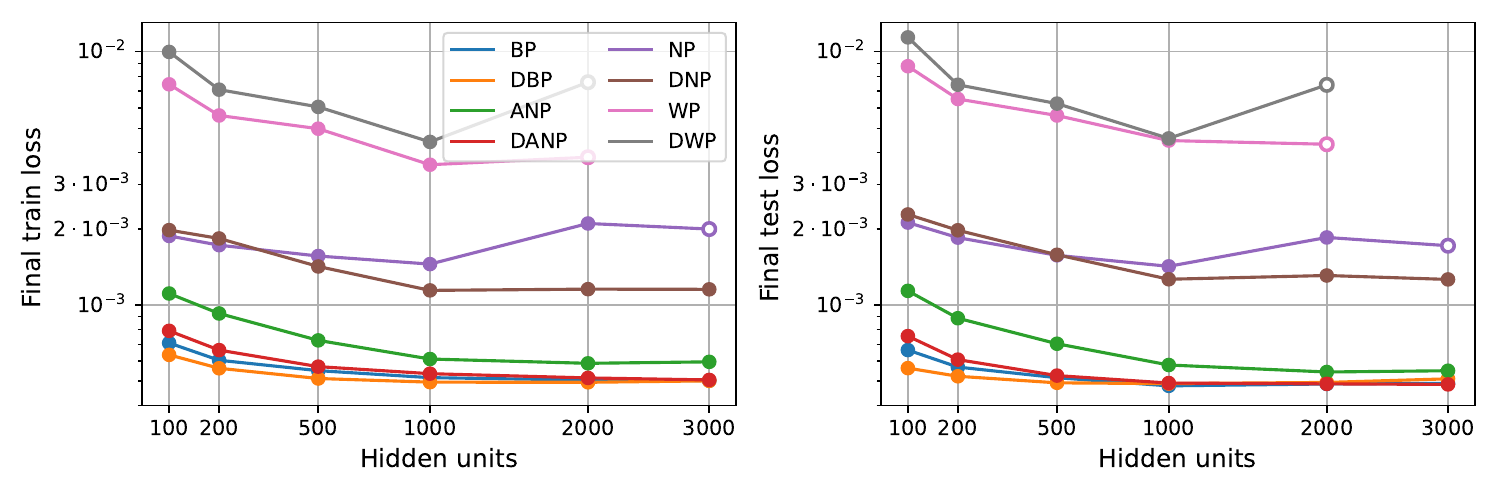}
    \caption{\textbf{Scalability.} Panels show final train and test loss. Each data point represents the mean of five runs. Solid circles denote stability across all runs, with the mean calculated from the five stable runs. Empty circles indicate instability in some runs, and the mean is computed solely from the remaining stable executions. The absence of data points indicates that all the runs were unstable.}
    \label{fig:scalability}
    \end{figure}
    
Figure~\ref{fig:scalability} shows the final performance over the train and test sets of networks with different configurations. Each configuration was repeated using five random seeds, and the results were computed as the mean of the successful executions within the set of five runs. 

The results of this experiment reveal two key findings.  Firstly, NP and WP exhibit inadequate scalability, leading to unstable runs or inferior performance when employed in the training of large networks. Secondly, ANP (and its variation incorporating the decorrelation mechanism, DANP) is the only perturbation-based method capable of effective scaling to larger networks, showing performance on par with BP and enhanced generalization when augmented with the decorrelation mechanism.

\subsection{Decorrelation results}
Here, we investigate the functionality of the decorrelation mechanism within the networks. Using the parameters that yield optimal task-dependent performance, we compare the degree of correlation in the neural outputs between a network incorporating the decorrelation mechanism and one that does not. To achieve this, we visualize the mean squared correlation across epochs during training on the 1-hour ahead weather prediction task.

\begin{figure}[!ht]
    \centering
    \includegraphics[width=0.87\textwidth]{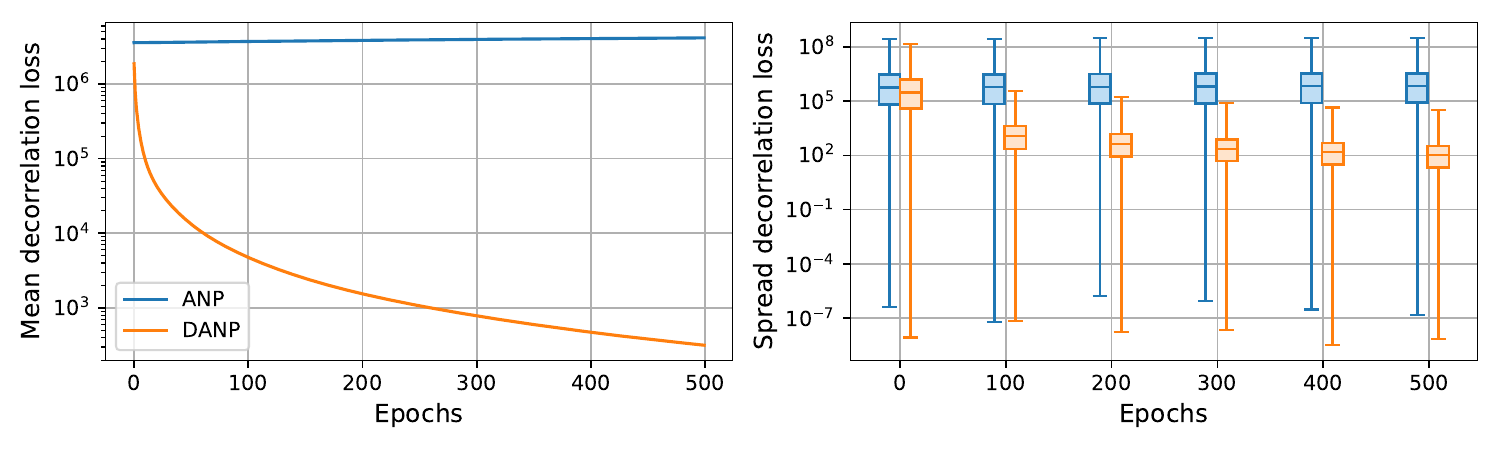}
    \caption{\textbf{Decorrelation loss during training.} The lower triangular correlation matrix is computed with the hidden units during the input presentation and subsequently squared and averaged across epochs. On the left, the mean decorrelation loss is depicted. On the right, a boxplot showing the spread of the decorrelation loss values is shown. Statistics were generated during training on the 1-hour ahead weather prediction task.}
    \label{fig:correlation_during_train}
    \end{figure}

Figure~\ref{fig:correlation_during_train} illustrates the degree of correlation within the hidden units. This represents the decorrelation loss, calculated as the mean squared off-diagonal values of the lower triangular covariance matrix, computed with the recurrent inputs, $x^*_t$, to the hidden units at each timestep during input presentation, across epochs. 


The outcomes of these experiments reveal a low degree of correlation among the hidden units in the network featuring the decorrelation mechanism compared to the network that does not incorporate it. However, as depicted in Figure~\ref{fig:correlation_during_train}, the parameters that yield optimal task-dependent performance do not lead to completely decorrelated neural outputs. These results suggest that a small level of correlation is needed to achieve optimal task-dependent performance, which may be due to the need for the weight updates to be able to keep up with decorrelation parameter updates.

\section{Discussion}

Backpropagation through time is the default algorithm to train RNNs. Like backpropagation, it relies on the computation and propagation of gradients for weight updates. However, the need to unroll the RNN over time makes this algorithm computationally demanding and memory-intensive, especially with long input sequences. Furthermore, the non-local nature of its updates and the requirement to backtrack through time once the input sequence concludes can pose some challenges in implementing this training method.
In this paper, we introduced a viable alternative for training RNNs. Instead of using explicit gradients, our method approximates the gradients stochastically via perturbation-based learning. To this end, we extended the decorrelated activity-based node perturbation approach~\citep{dalm2023effective}, which approximates stochastic gradient descent more accurately than other perturbation-based methods, to operate effectively in the time domain using RNNs. 


The results of our extensive validation show similar performance, convergence time and scalability of ANP and DANP when compared to BP. Remarkably, our approach exhibits superior performance, convergence time and scalability compared to standard perturbation-based methods, NP and WP. These promising findings suggest that perturbation-based learning holds the potential to rival BP's performance while retaining inherent advantages over gradient-based learning. Notably, the computational simplicity of our method stands out, especially compared to BP, which requires a specific second phase for error computation across networks and a dedicated circuit for error propagation. This simplicity is particularly advantageous for RNNs, eliminating the need for time-unrolling that significantly increases computational load. 
Additional experiments, detailed in Appendix \ref{appendix:comp_memory_costs}, empirically validate our claims that perturbation-based methods are computationally simpler and less memory-intensive than the standard gradient-based backpropagation method. Note that other gradient-free learning methods for RNNs have been proposed that do not require separate learning passes, such as the random feedback local online (RFLO) algorithm~\citep{murray2019local}. We show that our method still compares favourably in terms of learning efficiency and computational cost in Appendix \ref{appendix:comp_RFLO}.
In follow-up work, 
further validation on very large neural networks is necessary. This step is usually a common challenge in assessing the efficacy of alternative learning algorithms~\citep{bartunov2018assessing}. Other avenues for further research include exploring performance in more complex gated recurrent structures like LSTM and GRU units, as well as further validation on other challenging real-world datasets.

Our extensions differ from typical approaches in how the reinforcement signal is computed. While conventional extensions considers a global loss or aggregate local losses over time into a single loss per sequence and use it as the reinforcement signal, our methods directly incorporate local losses in time into the updates as the reinforcement signal. This allows for more informed and effective updates without increasing the computational load, considering peaks of high or low performance at specific time steps. Furthermore, both the clean and the noisy forward pass can be parallelized in this approach by using two identical copies of the model, making it compatible with online learning setups. This enables our synaptic rules to compute and implement updates online at every time step, reducing memory requirements as updates are applied immediately without storage. In this sense, our approach resembles the RTRL algorithm for computing gradients in RNNs in a forward manner~\citep{williams1989learning,zenke2020brain}. 
A limitation of our approach using local losses is its limited compatibility with delayed or sparse rewards over time. However, as demonstrated by our own results, as well as those of \citet{zenke2020brain} and \citet{Bellec2019}, local methods can still work well in such delayed settings.

Our approach has several limitations regarding biological fidelity. Our neural network models are significantly simplified, lacking spikes, structural plasticity, and detailed temporal dynamics and neural structure. However, our results are an important step in developing more biologically realistic learning approaches, while performing similarly to backpropagation, which is challenging to implement in biological circuits~\citep{lillicrap2020backpropagation, whittington2019theories}. Numerous methods have been proposed for learning synaptic weights in artificial neural networks as alternatives, such as feedback alignment~\citep{lillicrap2016random,nokland2016direct}, target propagation~\citep{bengio2014auto,lee2015difference, ahmad2020gait, bengio2020deriving}, dendritic error propagation~\citep{guerguiev2017towards, sacramento2018dendritic}, and spiking implementations combining global and local errors~\citep{Bellec2019}. However, these approaches generally still require computing parameter or neuron-specific errors, necessitating specific and complex circuits for error propagation. In contrast, our approach uses global errors akin to neuromodulatory signals in biological neural networks~\citep{schultz1998predictive, doya2002metalearning,marder2012neuromodulation, brzosko2019neuromodulation}. These global errors are uniformly spread across the network, eliminating the need for pathways to compute and deliver specific errors to each neuron. A key distinction of perturbation-based learning approaches, like ours, is their use of intrinsic brain noise for synaptic plasticity~\citep{faisal2008noise}, viewing noise as a beneficial feature rather than an obstacle. This mirrors the biological principle of using noise as a mechanism for learning, crucial for adapting to dynamic and unpredictable environments. Finally, our approach's active decorrelation method is similar to decorrelation in the brain, thought to be implemented through neural inhibition~\citep{ecker2010decorrelated, chini2022increase}. This process reduces correlations among neurons, promotes efficient information coding, enhances learning efficiency, and mirrors the adaptive processes found in biological neural networks.


While not fully biologically realistic, our models have sufficient fidelity to be highly relevant for neuromorphic computation. The local nature of the required computations, combined with a global learning signal, facilitates the deployment of these methods on neuromorphic hardware~\citep{Sandamirskaya2022,paredes2024fully}, devices that rely on distributed, localized processing units that mimic biological neurons. Embracing noise as a mechanism for learning can be highly suitable in settings where the computational substrate shows a high degree of noise~\citep{gokmen2021enabling}. This even holds when the noise cannot be measured since ANP still functions when we compare two noisy passes rather than a clean and a noisy pass~\citep{dalm2023effective}. This resilience to noise underscores the method’s suitability for real-world neuromorphic devices where noise is ubiquitous and difficult to control. Additionally, the gradient-free nature of the approach becomes valuable in settings where the computational graph contains non-differentiable components as in spiking recurrent neural networks; a type of network where effective training methods are still under exploration~\citep{tavanaei2019deep, wang2020supervised, neftci2019surrogate}.

Concluding, our present findings open the door to efficient gradient-free training of RNNs, offering exciting prospects for future research and applications in artificial intelligence, neuroscience and neuromorphic computing.

\bibliographystyle{apalike}
\bibliography{references}

\newpage 

\appendix

\section{Comparison of node perturbation methods}
\label{appendix:typicalNP_vs_ourNP}

\begin{figure}[!ht]
    \centering
    \includegraphics[width=\textwidth]{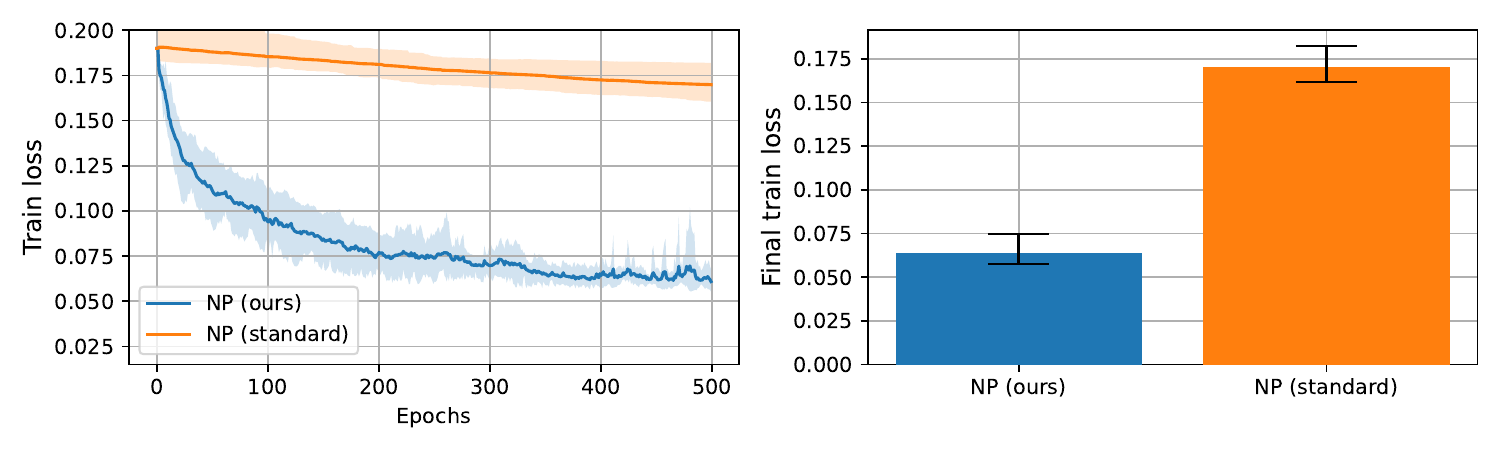}
    \caption{\textbf{Copying memory results for different implementations of NP.} On the left, the performance during training is depicted. On the right, the final performance is shown, calculated as the mean performance over the last 50 epochs.}
    \label{fig:copying_NPvsNP}
    \end{figure}

We here compare the conventional implementation of node perturbation, as in Equation~\ref{eq:np_conventional} with our implementation, as in Equation~\ref{eq:np_ours}.
To this end, we compare the performance of both approaches by solving the copying memory task, described in Section~\ref{copying_memory_task}, using the same setup as in our main experiments. As in the main experiments, five different executions with random seeds are carried out. The mean, maximum and minimum of these executions are then depicted. The results of this experiment are shown in Figure~\ref{fig:copying_NPvsNP}.
The experimental results indicate that our implementation significantly outperforms the typical node perturbation approach.

\section{Hyperparameter settings}
\label{appendix:hyperparameters}

In all experiments, certain hyperparameters remain fixed while others are fine-tuned to optimize the performance of each learning algorithm on specific datasets. This approach ensures a fair comparison between different algorithms. The number of epochs, hidden units, batch size, and Gaussian noise variance were kept consistent across learning algorithms for the various datasets, as detailed in Table~\ref{table:hyperparams}. 
The value selected for the Gaussian noise variance is the optimal value within the range $[10^{-1}, 10^{-4}]$. 

\renewcommand{\arraystretch}{1.5}
\begin{table}[h] 
\caption{Hyperparameters and their values used in the different experiments. Asterisks indicate that the value differs for each learning algorithm.}
\centering
\begin{small}
\begin{tabular}{ c p{4cm} c c c} 
\multirow{2}{*}{\textbf{Hyperparameter}} & \centering \multirow{2}{*}{\textbf{Description}} & \multicolumn{3}{c}{\textbf{Value per experiment}} \\ 
 \cline{3-5} 
 & & \textbf{Mackey-Glass} & \textbf{Copying} & \textbf{Weather} \\ 
 \hline
 $\sigma^2$ & Gaussian noise variance & $10^{-2}$ & $10^{-2}$ & $10^{-2}$\\
 \hline
 $\eta$ & Learning rate for  forward and recurrent weights & * & * & * \\
 \hline
 $\epsilon$ & Learning rate for  decorrelation weights & * & * & * \\
 \hline
 $N$ & Number of hidden units & 1000 & 500 & 1000 \\
 \hline
 batch & Batch size & 10 & 1 & 10 \\
 \hline
 epochs & Number of training epochs & 500 & 500 & 500 \\
 \hline
\end{tabular}
\end{small}
\label{table:hyperparams}
\end{table} 

In contrast, the learning rates $\eta$ and $\epsilon$ varied for each algorithm depending on the dataset, as detailed in Table~\ref{table:LRs}.
The values for the learning rates are the optimal values within the range $[10^{-1}, 10^{-10}]$. 
Learning rates exceeding the depicted values yielded unstable runs, characterized by weights becoming undefined due to exploding values, or optimization getting stuck in local minima, resulting in inferior final performance. Conversely, smaller values resulted in slower learning, leading to inferior final performance. The backpropagation algorithm was employed in combination with the Adam optimizer using the default parameters.

\begin{table}[h] 
\caption{Learning rates $\eta$ and $\epsilon$ for the different learning algorithms across datasets. Dashes indicate that the parameter is not used.}
\centering
\begin{small}
\begin{tabular}{l c c c c c c c c c} 
  & & \textbf{NP} & \textbf{DNP} & \textbf{WP} & \textbf{DWP} & \textbf{ANP} & \textbf{DANP} & \textbf{BP} & \textbf{DBP} \\
 \hline
 \multirow{2}{*}{\textbf{Mackey-Glass}} & $\eta$ & $5\mathrm{e}{-4}$ & $5\mathrm{e}{-4}$ & $5\mathrm{e}{-4}$ & $5\mathrm{e}{-4}$ & $3\mathrm{e}{-6}$ & $3\mathrm{e}{-5}$ & $5\mathrm{e}{-7}$ & $5\mathrm{e}{-7}$ \\ 
 \cline{2-10} 
 & $\epsilon$ & - & $5\mathrm{e}{-9}$ & - & $5\mathrm{e}{-9}$ & - & $5\mathrm{e}{-9}$ & - & $5\mathrm{e}{-7}$ \\
 \hline
  \multirow{2}{*}{\textbf{Copying}} & $\eta$ & $5\mathrm{e}{-2}$ & $5\mathrm{e}{-2}$ & $1\mathrm{e}{-3}$ & $1\mathrm{e}{-3}$ & $4\mathrm{e}{-5}$ & $4\mathrm{e}{-5}$ & $1\mathrm{e}{-4}$ & $1\mathrm{e}{-4}$ \\ 
 \cline{2-10} 
 & $\epsilon$ & - & $1\mathrm{e}{-4}$ & - & $1\mathrm{e}{-4}$ & - & $1\mathrm{e}{-4}$ & - & $1\mathrm{e}{-4}$ \\
 \hline
  \multirow{2}{*}{\textbf{Weather}} & $\eta$ & $1\mathrm{e}{-2}$ & $1\mathrm{e}{-2}$ & $1\mathrm{e}{-3}$ & $1\mathrm{e}{-3}$ & $3\mathrm{e}{-5}$ & $1.5\mathrm{e}{-4}$ & $5\mathrm{e}{-7}$ & $1\mathrm{e}{-6}$ \\ 
 \cline{2-10} 
 & $\epsilon$ & - & $1\mathrm{e}{-7}$ & - & $1\mathrm{e}{-7}$ & - & $5\mathrm{e}{-7}$ & - & $1\mathrm{e}{-7}$\\
 \hline
\end{tabular}
\end{small}
\label{table:LRs}
\end{table}

\section{Computational and memory costs analysis}\label{appendix:comp_memory_costs}

\begin{figure}[!ht]
    \centering
    \includegraphics[width=0.87\textwidth]{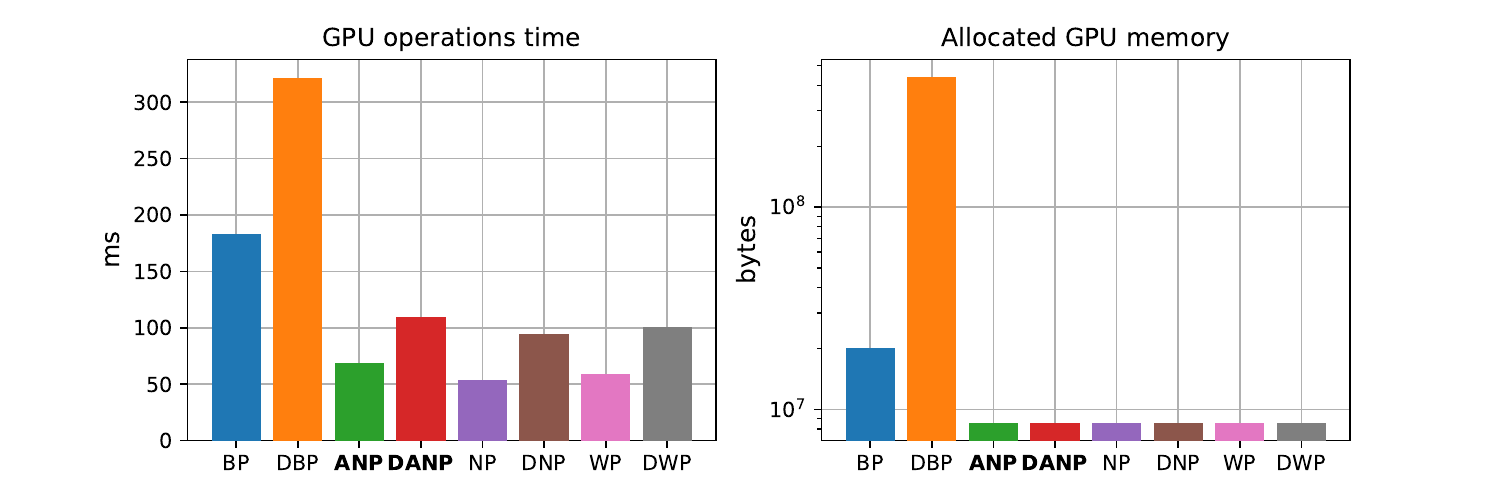}
    \caption{\textbf{Computational and memory costs.} Each data point represents the average of five executions. The analysis is conducted on a single sample from the copying memory dataset, presented only once. The left plot uses a normal scale, while the right plot employs a logarithmic scale for better visualization.}
    \label{fig:comp_memory_costs}
\end{figure}

Here, we profile the computational and memory costs of various learning algorithms considered in this paper. The analysis is conducted using the copying memory task, as it is the most studied and standard dataset in the literature. We measure the GPU operations time (in milliseconds) and allocated GPU memory (in bytes) for each algorithm on a single sample, presented one time, utilizing the PyTorch Profiler\footnote{ \url{https://pytorch.org/docs/stable/profiler.html}.}. To account for potential hardware and system variability, such as GPU load, background processes, and memory management, each algorithm is analyzed five times, and the results are averaged. The GPU used for this analysis is the Nvidia Tesla T4\footnote{ \url{https://images.nvidia.com/aem-dam/en-zz/Solutions/design-visualization/technologies/turing-architecture/NVIDIA-Turing-Architecture-Whitepaper.pdf}.}. The results of this analysis are presented in Figure \ref{fig:comp_memory_costs}.

Our empirical findings align with what is well-established in the literature: perturbation-based learning approaches are significantly less demanding in terms of computational and memory requirements compared to the standard gradient-based method, backpropagation. The GPU operations time is substantially lower for all perturbation-based methods considered in this study, with node perturbation being the most efficient, followed closely by weight perturbation and activity-based node perturbation. The allocated GPU memory remains comparable across perturbation-based methods, while it increases exponentially in backpropagation, particularly when incorporating the decorrelation mechanism. Unrolling the network over time with decorrelation steps results in substantial memory overhead, as all intermediate steps of all variables must be stored for each time step, unlike in perturbation-based methods.

\section{Comparison to another gradient-free method}\label{appendix:comp_RFLO}

To further extend the comparison of our approach with other gradient-free methods for training RNNs, we selected random feedback local online (RFLO). RFLO shares striking similarities with our approach and demonstrates highly efficient learning coupled with computational efficiency. Moreover, it employs eligibility traces and local updates in an online setup.

The setup for the RNN model used for the RFLO is the same as the one we employed, with the only difference being that the state of the hidden neurons evolves using a time constant $\tau$. This, while allowing for a more powerful temporal integration, adds computational complexity and leads to slower neural responses. The hidden units are defined as
\begin{equation}
    x_t = \left( \frac{1}{\tau} \right) f(A u_t + R x_{t-1}) + \left(1 - \frac{1}{\tau}\right) x_{t-1}.
\end{equation} 

The weight updates are defined as
\begin{align}
    \Delta A &= \varepsilon_t x_t,
\,,\quad    \Delta B = \left[ B \varepsilon_t \right] p_t,
\,,\quad    \Delta R = \left[ B \varepsilon_t \right] q_t,
\end{align}
where $B$ denotes fixed random feedback weights, $\varepsilon_t$ denotes a local error defined as $y_t - y^*_t$, and $p_t$ and $q_t$ are eligibility traces capturing previous neural states defined as
\begin{align}
    p_t &= \left( \frac{1}{\tau} \right) f^\prime (a_t) x_{t-1} + \left(1 - \frac{1}{\tau}\right) p_{t-1}
\\  q_t &= \left( \frac{1}{\tau} \right) f^\prime (a_t) u_{t-1} + \left(1 - \frac{1}{\tau}\right) q_{t-1}
\end{align}

In our experiments, the parameter $\tau$ is set to 10 as in the original work, and the learning rate is empirically selected to be the optimal value in the task within the range $[10^{-1}, \ldots, 10^{-10}]$, with $10^{-5}$ being the optimal value.

\begin{figure}[!ht]
\centering
\begin{flushleft}\:\: \:\: \:\: \:\: \textbf{a)}\end{flushleft} 
\includegraphics[width=0.8\textwidth]{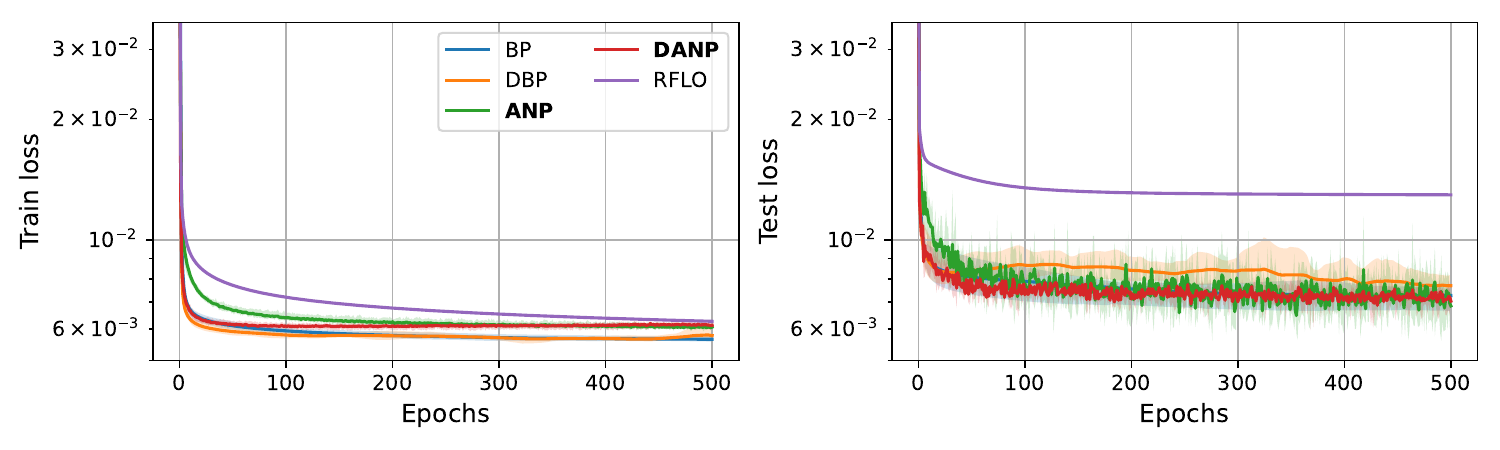}
\begin{flushleft}\:\: \:\: \:\: \:\: \textbf{b)}\end{flushleft} 
\includegraphics[width=0.8\textwidth]{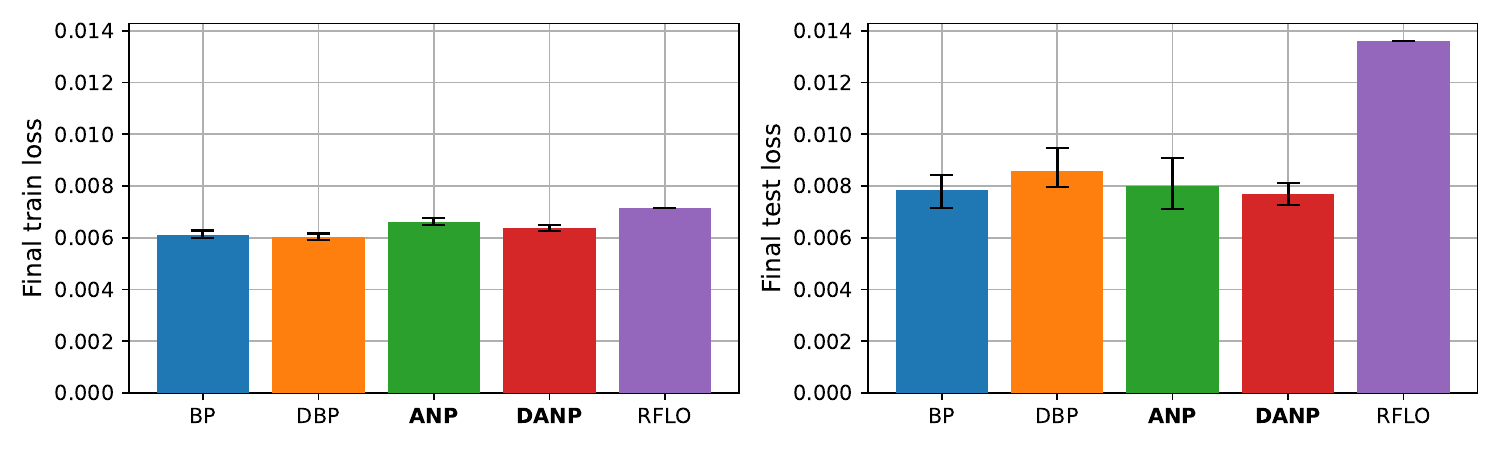}
\caption{\textbf{48-hour ahead weather prediction results including RFLO.} \textbf{a)} Performance during training over the train and test set for the different methods, represented in a logarithmic scale. \textbf{b)} Performance for the different methods, computed as the mean performance over the last 50 epochs.}
\label{fig:48h_weather_rflo}
\end{figure}

For simplicity, we focused on the most effective perturbation-based approaches, ANP and DANP, and the gradient-based baselines, BP and DBP. The learning efficiency comparison was conducted on the 48-hour weather prediction task, which is the most challenging task in our study. The results of this comparison are presented in Figure \ref{fig:48h_weather_rflo}.

We observe that RFLO learns the task efficiently but lags slightly behind other methods in terms of convergence speed. While its final training performance is comparable to the other approaches, its generalization capability (in terms of the test loss) is much lower.

\begin{figure}[!ht]
    \centering
    \includegraphics[width=0.87\textwidth]{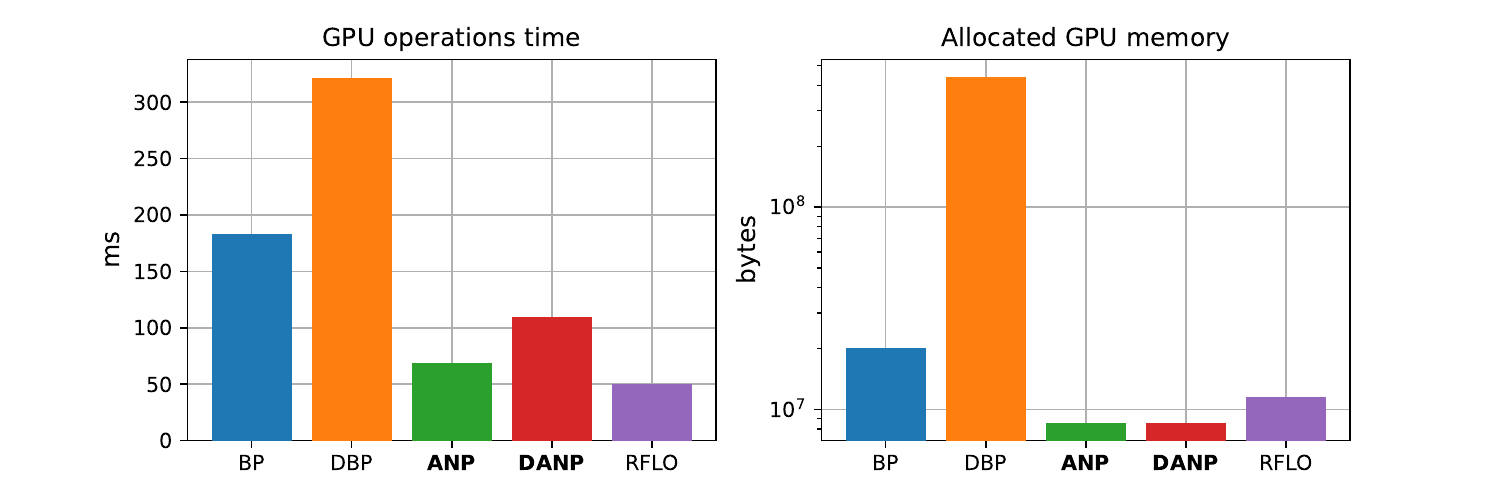}
    \caption{\textbf{Computational and memory costs.} Each data point represents the average of five executions. The analysis is conducted on a single sample from the copying memory dataset, presented only once. The left plot uses a normal scale, while the right plot employs a logarithmic scale for better visualization.}
    \label{fig:comp_memory_costs_RFLO}
\end{figure}

The computational cost analysis is performed following the procedure described in Section \ref{appendix:comp_memory_costs} of this document. The results of this analysis are shown in Figure \ref{fig:comp_memory_costs_RFLO}.

RFLO is slightly more efficient in terms of operation time but slightly less efficient in terms of memory usage.

\section{Additional weather prediction experiments}\label{appendix:extra_exp}

\begin{figure}[!ht]
    \centering
    \begin{flushleft}\:\: \textbf{a)}\end{flushleft}
    \includegraphics[width=0.88\textwidth]{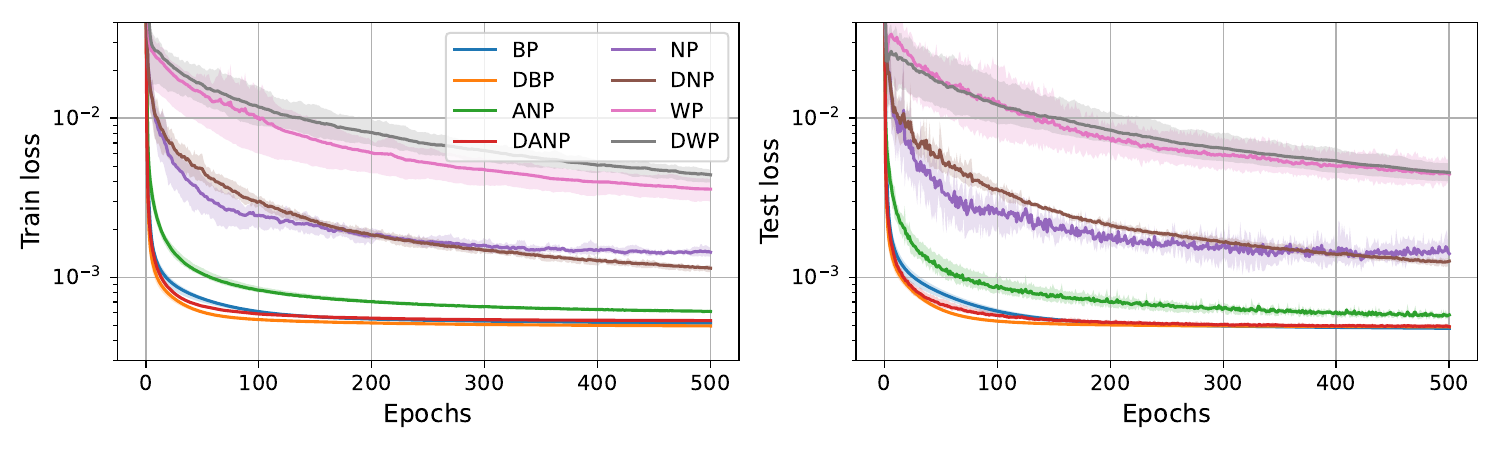}
    \begin{flushleft}\:\: \textbf{b)}\end{flushleft}
    \includegraphics[width=0.88\textwidth]{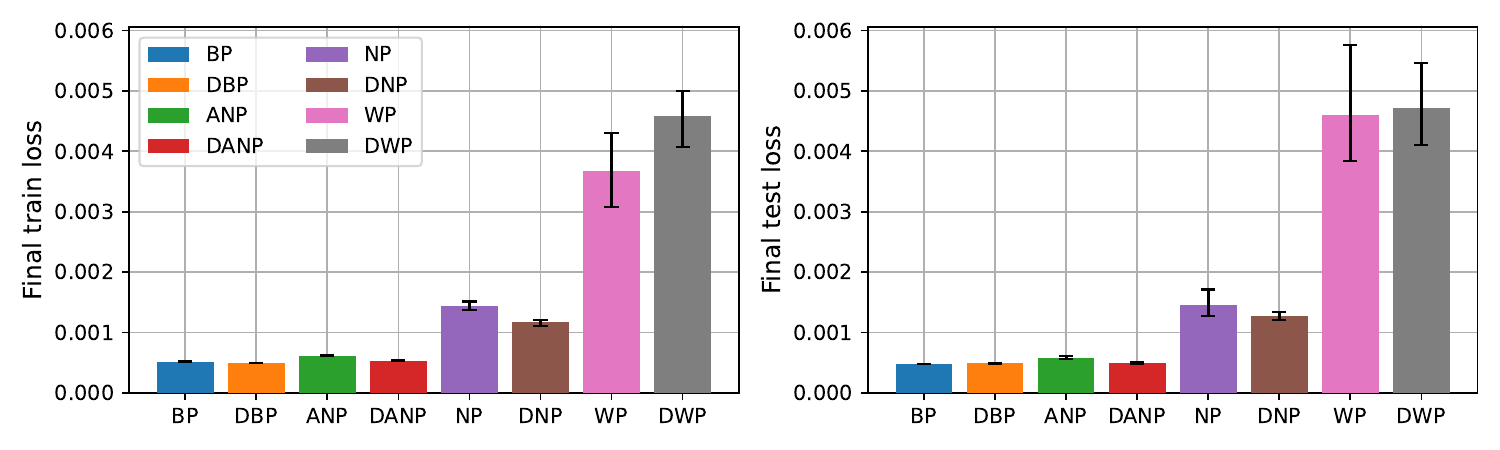}
    \caption{\textbf{1-hour ahead weather prediction results.} \textbf{a)} Performance during training over the train and test set for the different methods, represented in a logarithmic scale. \textbf{b)} Performance for the different methods, computed as the mean performance over the last 50 epochs.}
    \label{fig:1h_weather}
    \end{figure}

\begin{figure}[!ht]
    \centering
    \begin{flushleft}\:\: \textbf{a)}\end{flushleft}
    \includegraphics[width=0.88\textwidth]{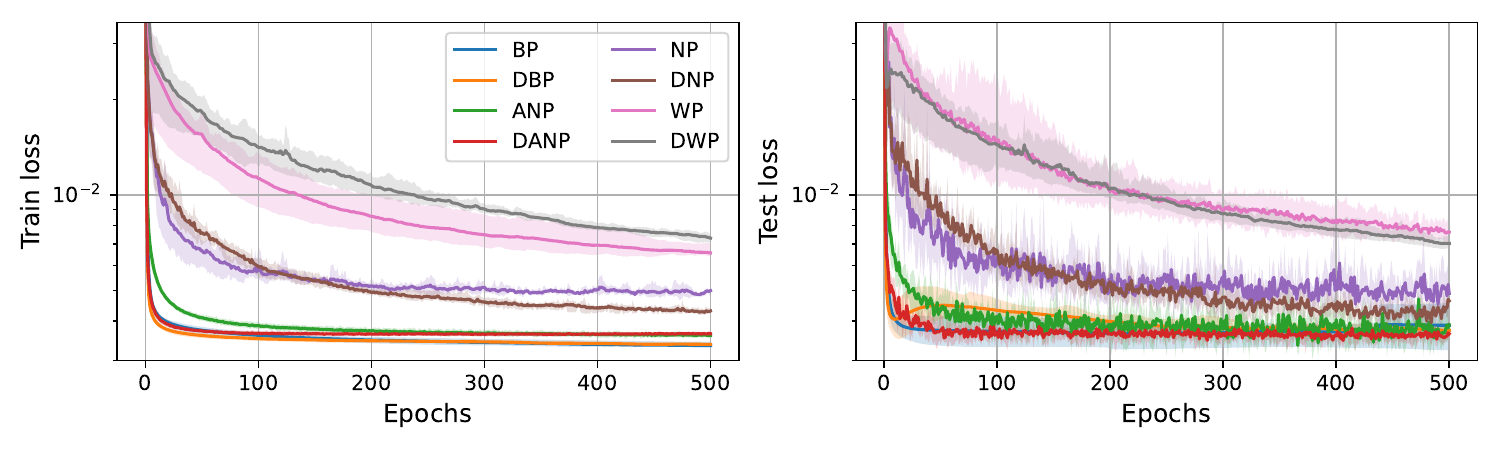}
    \begin{flushleft}\:\: \textbf{b)}\end{flushleft}
    \includegraphics[width=0.88\textwidth]{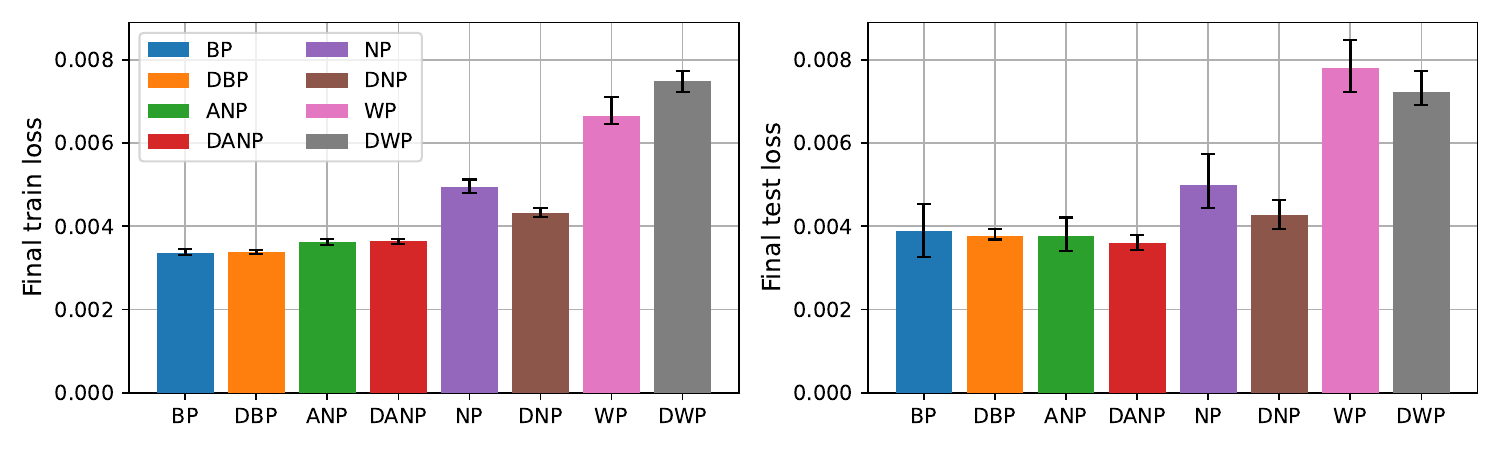}
    \caption{\textbf{24-hours ahead weather prediction results.} \textbf{a)} Performance during training over the train and test set for the different methods, represented in a logarithmic scale. \textbf{b)} Performance for the different methods, computed as the mean performance over the last 50 epochs.}
    \label{fig:24h_weather}
    \end{figure}

These experiments mirror the primary weather prediction experiments. However, in this case, the networks are tasked with predicting the target feature 1-hour and 24-hours ahead. See Figures~\ref{fig:1h_weather} and ~\ref{fig:24h_weather}, respectively.

\end{document}